\documentclass[11pt]{article}

\usepackage[final]{acl}

\usepackage{times}
\usepackage{latexsym}

\usepackage{bm}

\usepackage{booktabs}
\usepackage{tabularx}
\usepackage{array}
\usepackage{enumitem}
\usepackage{amsmath}
\usepackage{graphicx}
\usepackage{enumitem}
\usepackage{hyperref}

\usepackage{hyperref}


\usepackage[table]{xcolor}
\definecolor{lightgrayrow}{gray}{0.92}
\usepackage{booktabs}
\usepackage{multirow}
\usepackage{tabularx}
\usepackage{siunitx} 
\usepackage[T1]{fontenc}

\usepackage[utf8]{inputenc}

\usepackage[table]{xcolor}

\usepackage{pifont} 

\usepackage{amsmath}

\usepackage{graphicx}

\usepackage{microtype}

\usepackage{inconsolata}

\usepackage{graphicx}

\title{SCHEDBench: A Benchmark for Evaluating LLM Constraint Faithfulness in Natural-Language Combinatorial Scheduling}

\author{Shrenil Shaun Sharma\thanks{~~Equal contribution.} \\
  Independent Researcher\\
  San Francisco, CA, USA\\
  \texttt{shrenil19+research@gmail.com} \\\And
  Avi Sharma\footnotemark[1] \\
  Department of Electrical \\
  Engineering and Computer Sciences \\
  University of California, Berkeley \\
  \texttt{avi\_sharma@berkeley.edu} \\}

\begin{document}
\maketitle
\begin{abstract}
This paper introduces \textbf{SCHEDBench} \footnote{The Benchmark and Code are available at \url{https://huggingface.co/datasets/SCHEDBench/SCHEDBench}.}, a natural-language benchmark for evaluating  \textbf{combinatorial scheduling} constraint faithfulness under surface-form variation. Grounded in canonical scheduling instances and solver-derived feasibility and optimality, SCHEDBench assesses whether large language models (\textbf{LLMs}) generate schedules with the same constraint-feasible behavior across varied natural-language (NL) surface forms. SCHEDBench spans 1,132 instances across job-shop scheduling problems (JSP), single and multi-mode resource-constrained project scheduling problems (RCPSP), nurse rostering/scheduling, and curriculum timetabling problems of varying difficulty. Instances are templated into natural language problems using domain-specific templates, themed entities, lexical-syntactic template rephrasing, and constraint-level surface-form variation—with reference solutions verified for feasibility and objective optimality. Across thirteen frontier and open-weight LLMs, we find that models are not reliably invariant to semantically equivalent renderings of the same scheduling problem. Surface-form variation reduces feasibility and induces above-noise shifts in per-instance hard-constraint violations on matched instances. Among the tested isolated axes, constraint reordering yields the clearest above-noise sensitivity.

\end{abstract}

\section{Introduction}

Logical reasoning and discrete optimization have emerged as a central focus of modern large language model (\textbf{LLM}) research, driving a proliferation of new methods and benchmarks to evaluate and advance structured reasoning tasks. Combinatorial scheduling however, remains an underexplored yet critical frontier \cite{chen2025solverinformedrl,abgaryan_starjob_2025}, requiring models to reason over complex resource and precedence constraints across exponentially large search spaces, where small interpretation errors can render an entire solution infeasible. Beyond its computational difficulty, combinatorial scheduling assesses LLMs in their abilities to preserve formal constraints across linguistically varied but semantically equivalent problem descriptions. Practical natural-language interfaces to optimization may express the same scheduling instance through different entity names, constraint orderings, domain framings, or paraphrases. A reliable model should generate schedules with the same constraint-satisfaction behavior for each equivalent description of the same formal instance. Existing optimization benchmarks often evaluate a single formulation of each instance, leaving unclear whether model failures reflect limits in combinatorial search, instability in natural-language constraint interpretation, or both.

To address this gap we introduce \textbf{SCHEDBench}, a broad natural-language benchmark for evaluating combinatorial scheduling constraint faithfulness under surface-form variation. SCHEDBench draws 1,132 instances from canonical scheduling benchmarks, academic libraries, and competition datasets. We translate each structured instance into natural language using a deterministic template-based verbalization pipeline, introducing controlled variation in constraint ordering, thematic framing, entity naming, and lexical-syntactic rephrasing, while preserving the source problem’s constraints, objective, and combinatorial structure. We benchmark diverse LLMs by prompting each model to generate a complete schedule and evaluate outputs with domain-specific solvers. Layered ablations isolate which surface-form axes destabilize each model’s feasible-solution set, while measuring variation in feasibility, objective quality, and constraint satisfaction. We pair outputs by source instance across equivalent renderings and measure per-instance violation-rate changes, treating shifts that exceed the stochastic variability band as evidence that surface-form variation drives genuine model sensitivity, rather than rendering randomness. Across tested frontier and open-weight LLMs, we find that no model is fully invariant to tested surface-form variations: feasibility degrades under equivalent renderings, and per-instance violation rates shift beyond rendering noise. Concretely, our contributions are:

\begin{itemize}
  \item \textbf{SCHEDBench}, a 1,132-instance natural-language benchmark for evaluating constraint faithfulness across multiple scheduling domains and instances of varying difficulty.
  \item An empirical evaluation demonstrating surface-form variation degrades feasibility and shifts per-instance constraint satisfaction beyond a seed-noise floor for a subset of tested LLMs, with layered ablations identifying constraint reordering as the dominant causal axis.
  \item A template-based verbalization pipeline that converts canonical scheduling instances into natural-language problems with controlled surface-form variation preserving constraint, objective, and combinatorial structure.

\end{itemize}

\section{Related Works}
\providecommand{\cmark}{\(\surd\)}
\providecommand{\xmark}{\(\times\)}
\providecommand{\pmark}{\(\triangle\)}
\renewcommand{\cmark}{\textcolor{green!60!black}{\ding{51}}}
\renewcommand{\xmark}{\textcolor{red!70!black}{\ding{55}}}
\renewcommand{\pmark}{\textcolor{orange}{\(\blacktriangle\)}}

\begin{table*}[t]
\centering
\small
\setlength{\tabcolsep}{4pt}
\renewcommand{\arraystretch}{1.12}
\begin{tabular*}{\textwidth}{@{\extracolsep{\fill}}lcccccc}
\hline
\textbf{Benchmark} &
\shortstack{\textbf{Verbalized}\\\textbf{Input}} &
\shortstack{\textbf{Scheduling}\\\textbf{Focus}} &
\shortstack{\textbf{Multi-}\\\textbf{Family}} &
\shortstack{\textbf{NL}\\\textbf{Output}} &
\shortstack{\textbf{Evaluates}\\\textbf{Feasibility}} &
\shortstack{\textbf{Evaluates}\\\textbf{Optimality}} \\
\hline

ConstraintBench {\scriptsize\cite{tso_constraintbench_2026}}
    & \cmark & \xmark & \xmark & \cmark & \cmark & \cmark \\
IndusCP {\scriptsize\cite{shi_constraintllm_2025}}
    & \cmark & \xmark & \xmark & \xmark & \cmark & \cmark \\
NATURAL PLAN {\scriptsize\cite{zheng_natural_2024}}
    & \cmark & \xmark & \xmark & \cmark & \cmark & \xmark \\
NLCO {\scriptsize\cite{jiang_reasoning_2026}}
    & \cmark & \xmark & \cmark & \xmark & \cmark & \cmark \\
NL4Opt {\scriptsize\cite{ramamonjison_nl4opt_2023}}
    & \cmark & \xmark & \xmark & \xmark & \xmark & \xmark \\
PlanBench {\scriptsize\cite{valmeekam_planbench_2023}}
    & \cmark & \xmark & \xmark & \cmark & \cmark & \xmark \\
R-ConstraintBench {\scriptsize\cite{jain_r-constraintbench_2025}}
    & \cmark & \cmark & \xmark & \xmark & \cmark & \xmark \\
StarJob {\scriptsize\cite{abgaryan_starjob_2025}}
    & \xmark & \cmark & \xmark & \cmark & \cmark & \cmark \\
TCP {\scriptsize\cite{ding_tcp_2025}}
    & \cmark & \xmark & \xmark & \cmark & \cmark & \xmark \\
ZebraLogic {\scriptsize\cite{lin_zebralogic_2025}}
    & \cmark & \xmark & \xmark & \cmark & \cmark & \xmark \\
\hline
\textbf{SCHEDBench}  & \cmark & \cmark & \cmark & \cmark & \cmark & \cmark \\

\hline
\end{tabular*}
\caption{
\textbf{Comparison of LLM benchmarks across six criteria relevant to constraint-faithful natural-language scheduling.}
(1)~\textit{Verbalized Input}: instances are expressed in natural language;
(2)~\textit{Scheduling Focus}: benchmark centers on scheduling;
(3)~\textit{Multi-Family}: covers multiple scheduling families;
(4)~\textit{NL Output}: models generate schedules in natural language;
(5)~\textit{Feasibility}: outputs are validated through programmatic constraint checking;
(6)~\textit{Optimality}: solutions are scored against optimal or best-known references.
}
\label{tab:benchmark-comparison}
\end{table*}

Evaluating language models on structured reasoning tasks remains a sustained focus in NLP, with benchmarks spanning natural language inference \cite{bowman_large_2015}, commonsense reasoning \cite{talmor_commonsenseqa_2019}, formal logical reasoning \cite{han_folio_2024} and code generation \cite{chen_evaluating_2021}, revealing both the breadth of LLM capabilities and systematic failure modes under formally specified tasks.

\subsection{LLM Robustness to Surface-Form Variation}
A consistent finding in NLP is that model behavior can be sensitive to surface form and prompt formatting in ways not predicted by task semantics alone. \citet{ribeiro_beyond_2020} introduced behavioral testing via controlled linguistic perturbations, demonstrating systematic failure under surface changes that preserve semantics. \citet{sinha_unnatural_2021} showed that transformer models are largely insensitive to word order, suggesting reliance on lexical, rather than structural cues. \citet{mccoy_right_2019} demonstrated that NLI models exploit shallow syntactic heuristics rather than logical form, collapsing under minimal surface perturbations that preserve meaning. \citet{sclar_quantifying_2024} extend this analysis to instruction-tuned LLMs, quantifying accuracy shifts under superficial reformulations across diverse tasks. SCHEDBench examines whether this brittleness extends to formal constraint structure in a domain where ground truth is solver-verifiable.

\subsection{Planning, Temporal, and Constraint-Based Reasoning}
PlanBench \cite{valmeekam_planbench_2023} establishes a baseline for goal-directed deterministic planning. NATURAL PLAN \cite{zheng_natural_2024} evaluates natural-language planning across trip, meeting, and calendar scheduling; TIMEBENCH \cite{chu_timebench_2024} and Test of Time \cite{fatemi_test_2024} probe ordering, duration, and arithmetic over time, the latter using synthetic construction to decouple temporal reasoning from factual recall. TCP \cite{ding_tcp_2025} extends this to constraint-based temporal planning with time zones and dynamic unavailability. These benchmarks advance structured temporal reasoning evaluation, including naturalistic dialogue settings with interdependent constraints. However, they primarily target temporal planning rather than scheduling families which contain resource disjunctions, precedence networks, and penalty-based scoring.

\subsection{Constraint Satisfaction and Combinatorial Optimization}
LLM evaluation on constraint satisfaction and combinatorial search has attracted considerable recent attention. ZebraLogic \cite{lin_zebralogic_2025} frames satisfiability problems as natural-language puzzles, evaluating whether models identify valid assignments. IndusCP \cite{shi_constraintllm_2025}, ConstraintBench \cite{tso_constraintbench_2026}, and NLCO \cite{jiang_reasoning_2026} broaden to constraint-programming and operations-research domains, with ConstraintBench evaluating solution quality against solver-verified references. Yet, none isolate scheduling-specific structures such as precedence chains, disjunctive machine contention, or soft-penalty tradeoffs.

\subsection{Scheduling-Specific LLM Evaluation}
Direct evaluation on scheduling problems remains sparse. StarJob \cite{abgaryan_starjob_2025} evaluates models only on job-shop scheduling and uses synthetic instances that do not reflect the combinatorial depth of canonical benchmarks. R-ConstraintBench \cite{jain_r-constraintbench_2025} similarly restricts evaluation to RCPSP with synthetic generation. SCHEDBench unifies JSP, RCPSP, timetabling, and nurse rostering under a common natural-language interface with systematic surface-form variation across each family.

\subsection{Natural Language to Formal Optimization}
NL4Opt \cite{ramamonjison_nl4opt_2023} evaluates LLMs as semantic parsers for optimization on linear programming word problems, with subsequent work extending to integer, mixed-integer, and broader mathematical programming settings \cite{huang_mamo_2024, yang_optibench_2025}. Whereas these benchmarks assess whether a produced formal representation is faithful to the source text, SCHEDBench evaluates whether models generate correct schedules end-to-end, using solvers for validation rather than as the required output format.

\section{SCHEDBench Benchmark Creation}

\subsection{Instance Curation}

To cover diverse scheduling task families, we construct instances from multiple canonical operations-research benchmarks. These problems fall into two primary categories: makespan minimization and penalty minimization.

\paragraph{Makespan Minimization Problems}
In Resource Constrained Project Scheduling Problems (RCPSP) and JSP, the objective is to minimize the elapsed time between the start of the first task and the completion of the last. In single-mode (SM), each activity has a fixed duration, consumes resources while executing, and must respect precedence constraints; in multi-mode (MM), each activity may be executed in one of several modes, each with a distinct resource consumption profile. We draw RCPSP instances from PSPLib \cite{kolisch_psplib_1997}, hereafter referred to as RCPSP-Lib. In JSP, each job consists of a fixed-order sequence of operations, each requiring exclusive machine access for a fixed duration. We draw all instances from the JSPLib Repository, a collection of nine established instance families detailed in Appendix~\ref{sec:appendixA-instance-families}, which cover a range of sizes.

\paragraph{Penalty Minimization Problems}
Unlike makespan-minimization problems, this category optimizes a weighted sum of constraint violations. For Curriculum Timetabling problems, we use the ITC-2007 Track 3 benchmark \cite{digaspero2007itc2007track3,bonutti2012benchmarking}, based on realistic university timetabling scenarios. For Nurse Rostering problems, we use the first International Nurse Rostering Competition (INRC-I) and the Nurse Scheduling Problems Library (NSPLib). Both assign nurses to shifts subject to coverage requirements and regulatory constraints, minimizing weighted penalty scores. NSPLib \cite{vanhoucke2007nsplib,vanhoucke2009characterization} provides instances parameterized by nurse count, shift types, and scheduling horizon, whereas INRC-I \cite{haspeslagh2014inrc} extends this with richer constraint sets and a broader range of horizon lengths.

\begin{figure}[t]
    \centering
    \includegraphics[width=\linewidth]{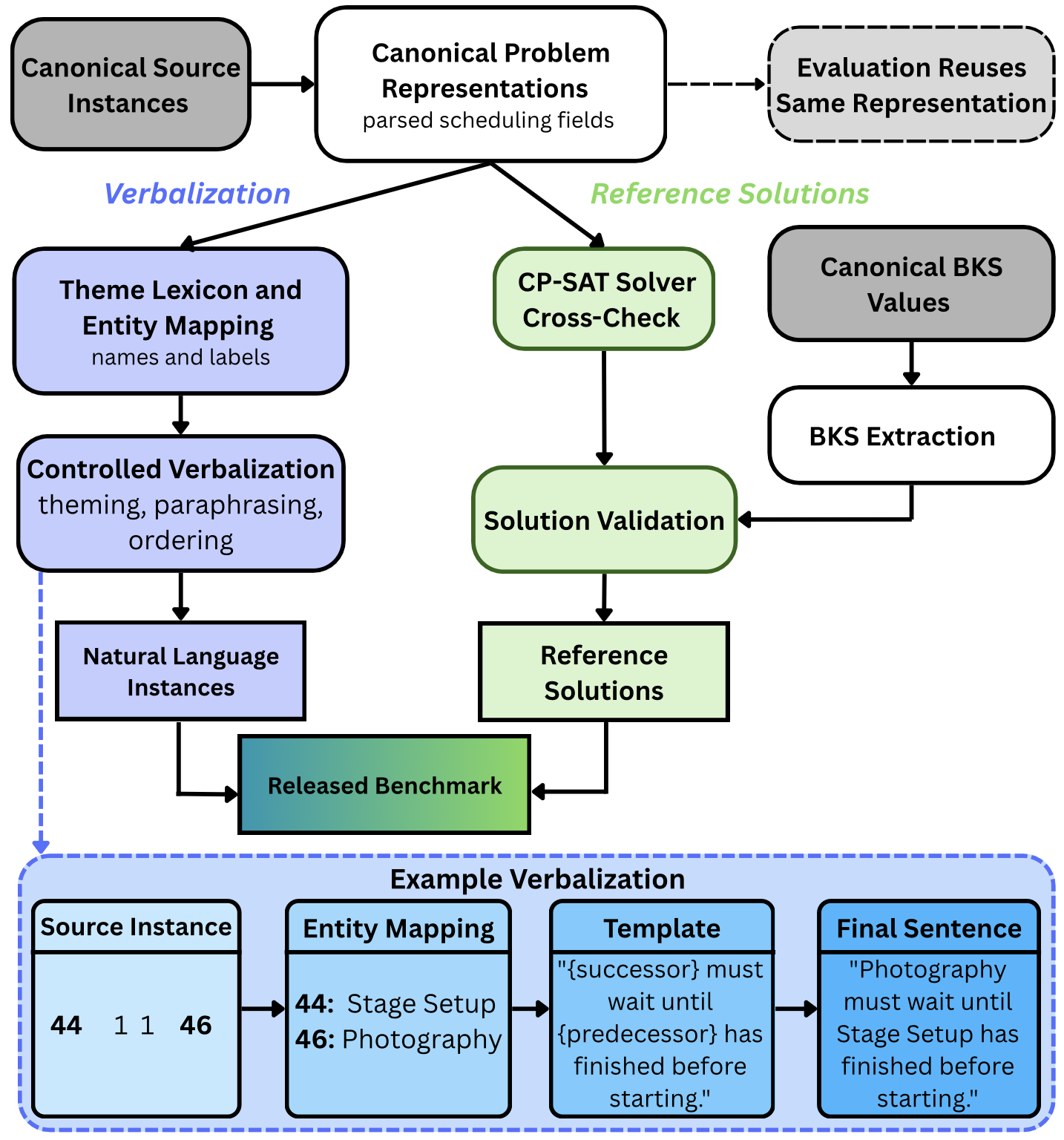}
    \caption{Overview of the SCHEDBench generation pipeline.}
    \label{fig:pipeline}
\end{figure}

\subsection{Data to Text Verbalization}

We verbalize each source instance through the controlled slot-filling pipeline shown in Fig\ref{fig:pipeline} rather than LLM-based generation. Each instance is parsed into structured constraint entries covering operations, resource capacities, precedence relations, coverage requirements, and penalty terms; verbalized with family-specific templates parsed from the source record. Entity substitutions are drawn from theme-assigned lexicons. Variation is controlled along three independently seeded axes: template selection, domain-theme assignment, and constraint re-ordering, adjusted to produce the ablations in Section 4. Verbalized instances include a family-specific background prompt with domain rules and shared assumptions. We manually audit 100 stratified instances for correctness; details appear in Appendix~\ref{sec:appendixA-faithfulness-audit}.

\paragraph{JSPLib}
Instances contain three constraint categories: situation framing, situation closing, and per-operation constraints. Job and location identifiers are drawn from pools with no inherent ordering, preventing entity co-occurrence shortcuts. Three job-level discourse structures vary surface form while preserving operation order: stepwise (independent clauses), chunked (range expressions), and ordered (semicolon lists), assigned per job.

\paragraph{RCPSP-Lib}
Instance generation follows a fixed semantic hierarchy: global resource capacities, activity definitions, and precedence statements. Precedence constraints are verbalized through causal, temporal, and rule-based formulations that vary surface syntax while preserving the underlying directed relation. Multi-Mode RCPSP extends this layout by lexically distinguishing renewable and nonrenewable resource pools and introducing explicit mode-selection sentences.

\paragraph{ITC}
Instances span seven constraint categories covering global counts, course definitions, room capacities, curriculum groups, and unavailability windows, realized across themes such as film production and university administration. Separate template banks handle single-, dual-, and multi-course curricula, expressing mutual exclusion at the group level rather than through implied surface wording.

\paragraph{INRC}
Instances are verbalized through categories covering skill definitions, shift definitions, contract rules, and coverage requirements. Shift definitions are routed through separate templates according to whether they require no skill, one skill, or multiple skills. Hard constraints use deontic modals (must, is required), while soft constraints carry evaluative language (is penalized for, should avoid) alongside explicit numerical weights.

\paragraph{NSPLib}
Instances span semantic categories organized into five blocks: base count declarations, shift definitions, coverage requirements, assignment costs, and paired case rules. The block structure enforces a fixed presentation order, ensuring all global parameters appear before per-worker and per-shift constraints. Coverage requirements are stated as exact staffing targets rather than minimum bounds.

\subsection{Reference Solutions}

Best-Known Solution (BKS) values are extracted from the same canonical benchmark repositories used to source the instances, corresponding either to proven optima or to best-known objective values reported in the original benchmark resources. To verify these references, we instantiate a family-specific problem structure for each source instance and solve it directly with OR-Tools CP-SAT under fixed budgets detailed in Appendix~\ref{sec:appendixA-bks-source-documentation}.

\begin{table*}[t]
\centering
\small
\setlength{\tabcolsep}{5pt}
\renewcommand{\arraystretch}{1.05}

\begin{tabular}{l l l c | c c c c}
\hline
\textbf{Source Family} &
\textbf{Objective} &
\textbf{Problem Type} &
\textbf{Mean Constraint Count} &
\textbf{Easy} &
\textbf{Medium} &
\textbf{Hard} &
\textbf{Total} \\
\hline

JSPLib     & Makespan            & JSP              & 695.5 & 139 & 43  & 80  & 262 \\
RCPSP-Lib  & Makespan            & RCPSP (SM)        & 321.8 & 100 & 100 & 200 & 400 \\
RCPSP-Lib  & Makespan            & RCPSP (MM)        & 740.0 &  100 & 100 & 100 & 300 \\
ITC        & Penalty Violations  & Timetabling       & 157.0 & 7   & 7   & 7   & 21 \\
INRC       & Penalty Violations  & Nurse Rostering   & 107.0 & 33  & 18  & 18  & 69 \\
NSPLib     & Penalty Violations  & Nurse Rostering   & 29.9 & 27  & 27  & 26  & 80 \\
\hline

\textbf{Total} & -- & -- & -- &
\textbf{406} &
\textbf{295} &
\textbf{431} &
\textbf{1132} \\
\hline
\end{tabular}

\caption{Composition of SCHEDBench across canonical source families, optimization objectives, problem domains, and source-specific difficulty partitions. Example of complete instance provided in Appendix Fig \ref{fig:schedbench-instance-example}.}
\label{tab:schedbench_full}
\end{table*}

\subsection{Benchmark Composition}

SCHEDBench comprises 1,132 instances drawn from six canonical scheduling source families: JSPLib, RCPSP-Lib SM, RCPSP-Lib MM, ITC, INRC, NSPLib. Each instance consists of a verbalized NL problem description paired with a family-specific background prompt; as described in Section 3.2, this prompt encodes shared domain context and optionally specifies the expected output format. Instances are classified into three difficulty tiers (Easy: 406, Medium: 295, Hard: 431), inherited from canonical sources or assigned using established optimality gaps (see Appendix ~\ref{sec:appendixA-difficulty}). Mean constraint counts range from 29.9 (NSPLib) to 740.0 (RCPSP-Lib MM), capturing substantial variation in problem complexity across families. Table 2 reports the full family-level breakdown, including problem types, optimization objectives, and difficulty distributions. We cite original benchmark papers, preserving instance identifiers, and release the pipeline under MIT and the derived benchmark under CC-BY 4.0; licensing details appear in Appendix~\ref{sec:appendixA-license-documentation}.

\section{Experimental Evaluation}

We evaluate open-weight and proprietary LLMs of varying sizes, on their ability to produce feasible, objective-minimizing solutions for SCHEDBench instances.

\begin{table*}[!t]
\centering
\small
\setlength{\tabcolsep}{3pt}
\sisetup{table-format=3.1, table-number-alignment=center}
\providecommand{\colhead}[1]{}%
\renewcommand{\colhead}[1]{{\scriptsize\bfseries\shortstack[c]{#1}}}%
\resizebox{\textwidth}{!}{%
\begin{tabular}{ll c *{13}{S} >{\columncolor{lightgrayrow}}S}
\toprule
& & & \multicolumn{13}{c}{\textbf{Feasibility rate (\%)}} & \multicolumn{1}{c}{} \\
\cmidrule(lr){4-16}
\textbf{Family} & \textbf{Diff.} & {$n_d$}
& {\colhead{gpt\\5.5}}
& {\colhead{gemini-3\\flash-preview}}
& {\colhead{qwen\\3.5-27b}}
& {\colhead{qwen\\3.5-397b}}
& {\colhead{qwen\\3.5-122b}}
& {\colhead{claude\\sonnet-4-6}}
& {\colhead{claude\\opus-4-6}}
& {\colhead{claude\\haiku-4-5}}
& {\colhead{llama\\3.3-70b}}
& {\colhead{llama-4\\maverick}}
& {\colhead{gpt-5.4\\mini}}
& {\colhead{gemini-3.1\\flash-lite}}
& {\colhead{gpt\\5.4}}
& {\textbf{Avg}} \\
\midrule
\multirow{3}{*}{ITC}
  & E &   7 &  14.3 &   0.0 &   0.0 &   0.0 &   0.0 &  28.6 &   0.0 &   0.0 &   0.0 &   0.0 &   0.0 &   0.0 &   0.0 &   3.3 \\
  & M &   7 &   0.0 &   0.0 &   0.0 &   0.0 &   0.0 &   0.0 &   0.0 &   0.0 &   0.0 &   0.0 &   0.0 &   0.0 &   0.0 &   0.0 \\
  & H &   7 &   0.0 &   0.0 &   0.0 &   0.0 &   0.0 &   0.0 &   0.0 &   0.0 &   0.0 &   0.0 &   0.0 &   0.0 &   0.0 &   0.0 \\
\midrule
\multirow{3}{*}{INRC}
  & E &  33 &  97.0 &  87.9 & 100.0 & 100.0 &  90.9 & 100.0 & 100.0 & 100.0 &  97.0 &  97.0 & 100.0 &  90.9 & 100.0 &  97.0 \\
  & M &  18 &  66.7 &  66.7 &  61.1 &  66.7 &  55.6 &  66.7 &  66.7 &  66.7 &  66.7 &  66.7 &  66.7 &  61.1 &  33.3 &  62.4 \\
  & H &  18 &   0.0 &   0.0 &   0.0 &   0.0 &   0.0 &   0.0 &   0.0 &  16.7 &   0.0 &   0.0 &   5.6 &   0.0 &   0.0 &   1.7 \\
\midrule
\multirow{3}{*}{NSPLib}
  & E &  27 & 100.0 &  74.1 &  59.3 &  48.1 &  59.3 &  22.2 &   0.0 &   0.0 &   0.0 &   0.0 &   0.0 &   0.0 &   0.0 &  27.9 \\
  & M &  27 &  85.2 &  92.6 &  66.7 &  63.0 &  74.1 &  22.2 &   0.0 &   0.0 &   0.0 &   0.0 &   0.0 &   0.0 &   0.0 &  31.1 \\
  & H &  26 &  88.5 &  88.5 &  53.8 &  69.2 &  53.8 &   7.7 &  11.5 &   0.0 &   0.0 &   0.0 &   0.0 &   0.0 &   0.0 &  28.7 \\
\midrule
\multirow{3}{*}{JSPLib}
  & E & 139 &  96.4 &  17.3 &  42.4 &  23.0 &   5.0 &   2.9 &   0.0 &   0.0 &   0.0 &   0.0 &   0.0 &   0.0 &   0.0 &  14.4 \\
  & M &  43 & 100.0 &   2.3 &  44.2 &  37.2 &   4.7 &   2.3 &   0.0 &   0.0 &   0.0 &   0.0 &   0.0 &   0.0 &   0.0 &  14.7 \\
  & H &  80 & 100.0 &   0.0 &  52.5 &  27.5 &   1.2 &   7.5 &   0.0 &   0.0 &   0.0 &   0.0 &   0.0 &   0.0 &   0.0 &  14.5 \\
\midrule
\multirow{3}{*}{RCPSP-MM}
  & E & 100 &  82.0 &  83.0 &  54.0 &  68.0 &  48.0 &  26.0 &  33.0 &   8.0 &  11.0 &   7.0 &   3.0 &   3.0 &   3.0 &  33.0 \\
  & M & 100 &  57.0 &  51.0 &   8.0 &  22.0 &   7.0 &  24.0 &   7.0 &   0.0 &   1.0 &   0.0 &   0.0 &   0.0 &   0.0 &  13.6 \\
  & H & 100 &  31.0 &  15.0 &   2.0 &   1.0 &   0.0 &  21.0 &   5.0 &   0.0 &   0.0 &   0.0 &   0.0 &   0.0 &   0.0 &   5.8 \\
\midrule
\multirow{3}{*}{RCPSP-SM}
  & E & 100 &  57.0 &  46.0 &   7.0 &  18.0 &   7.0 &   0.0 &   7.0 &   0.0 &   0.0 &   0.0 &   0.0 &   0.0 &   0.0 &  10.9 \\
  & M & 100 &  23.0 &   5.0 &   0.0 &   3.0 &   0.0 &   2.0 &   1.0 &   0.0 &   0.0 &   0.0 &   0.0 &   0.0 &   0.0 &   2.6 \\
  & H & 200 &   4.0 &   0.0 &   0.0 &   0.0 &   0.0 &   0.0 &   0.0 &   0.0 &   0.0 &   0.0 &   0.0 &   0.0 &   0.0 &   0.3 \\
\midrule
\rowcolor{lightgrayrow}
\multicolumn{2}{l}{\textbf{All}} & {\textbf{1132}}
& \bfseries 55.9 & 29.5 & 25.0 & 24.3 & 14.3 & 12.8 & 8.9 & 4.9 & 4.9 & 4.5 & 4.3 & 3.9 & 3.7 & \textbf{15.2} \\
\bottomrule
\end{tabular}%
}
\caption{Feasibility rate on the SCHEDBench-main evaluation set under full-variation (v4). Each family is split into Easy (E), Medium (M), and Hard (H), with counts $n_d$ in column 3 ($n=1132$ total). All model cells are feasibility rates as a percentage. The rightmost \textbf{Avg} column averages feasibility across models per bucket; the highlighted \textbf{All} row at the bottom gives each model's overall feasibility across all 1132 instances. Models are ordered left to right by descending overall feasibility.}
\label{tab:v4-difficulty}
\end{table*}

\subsection{Experiment Set Up}

\textbf{Models:} Selected open-weight models include Llama 3.3, Maverick 4, and Qwen 3.5, while proprietary models include GPT 5.4-mini, GPT 5.4, GPT 5.5, Claude Haiku 4.5, Claude Sonnet 4.6, Claude Opus 4.6, Gemini 3.1 Flash Lite, and Gemini 3.1 Flash Preview. Models are generally evaluated using their default inference settings to reflect standard ‘out-of-the-box’ performance.\footnote{Appendix~\ref{sec:appendixB} provides full model, inference, compute, usage, and evaluation-scale details.}

\vspace{0.5em}
\noindent\textbf{Inference:} Each instance is provided in a zero-shot setting: the LLM receives the problem instance, a family-specific prompt specifying the scheduling rules and output format, and a global system prompt requiring it to output only the final schedule without external tools or solvers. We use greedy decoding, to obtain deterministic outputs for each model-instance across repeated runs and adequate output limits to prevent truncation.

\subsection{Evaluation Sets}

We evaluate the full SCHEDBench-main set of 1,132 instances under full surface variation, serving as the primary benchmark evaluation. To isolate the effect of each surface-form variation, we construct targeted ablation sets over a matched 470-instance subset, enabling paired comparisons across direct canonical-to-NL translation (v1), constraint order variation (v2), thematic-domain variation (v3A), paraphrasing via lexical-syntactic rephrasing (v3B), and the combined full variation rendering (v4), holding the underlying formal instances fixed. The subset is stratified by subdomain, including all smaller families and fixed-size samples from larger families.

\subsection{Seed Sensitivity}
\label{sec:seed}
To verify that main or ablation-level results are not artifacts of the particular rendering seed used to control constraint order shuffling, theme assignment, or paraphrase selection, we conduct a seed-sensitivity ablation estimating a per-model noise floor for each randomized condition set v2, v3A, v3B respectively. For each condition $c \in \{\mathrm{v2}, \mathrm{v3A}, \mathrm{v3B}\}$
we generate $k=3$ independently seeded, domain-stratified renderings of
a fixed 150-instance subset: the rendering inherited from the
470-instance ablation set, plus two with new seeds. For model $m$,
condition $c$, seed pair $(s,s')$, and instance $i$, let
$\delta_{m,c}^{(s,s')}(i) = \lvert \mathrm{VR}_{m,c,s}(i) -
\mathrm{VR}_{m,c,s'}(i) \rvert$, the absolute per-instance change in
violation rate between seeds. The seed band is the distribution of
$\delta$ pooled over all three seed pairs:
\[
B_{m,c} = \bigcup_{(s,s')} \{\, \delta_{m,c}^{(s,s')}(i) : i = 1,\dots,N \,\}.
\]
We summarize each seed band by its median and interquartile range. A v1$\rightarrow$axis effect is judged beyond noise when the 95\%
bootstrap CI on $\mathrm{median}(\lvert\Delta\mathrm{VR}\rvert) -
\mathrm{median}(B_{m,c})$ lies above zero. Violation rate is used throughout, as feasibility is floor-bound on most families and provides no
seed-to-seed signal in those instances. For v3A/v3B, reseeding reapplies the same axis, so within-noise results indicate indistinguishability from rendering resampling not absence of effect. The seed test indicates whether an axis effect exceeds reseeding noise, but not how broadly the effect is distributed, therefore we report the full $|\Delta \mathrm{VR}|$ distribution, summarized by median and IQR.

\subsection{Metrics}

We score model outputs with a parser and family-specific verifier, after minimal post-processing removes non-substantive extraneous text, extracting the generated schedule, and verifying feasibility of all hard constraints.  

\vspace{0.5em}
\noindent{\textbf{SCHEDBench-Main:} Our primary metric is feasibility rate: the proportion of instances for which the model produces a valid schedule. We report feasibility both overall and by scheduling family, difficulty, and surface-form variant. For feasible schedules, we additionally measure objective quality using the percentage optimality gap. All SCHEDBench objectives are minimization (makespan, soft-constraint penalty), with lower being better. For an instance with model objective value $z$ and reference value $z^*$, the gap is computed as $(z - z^*) / z^* × 100$, where $z^*$ is the best-known objective value from the source benchmark library. Gap statistics are computed over feasible outputs only, since infeasible schedules do not define valid objective values. We also report median gap as a more robust summary when gaps contain large outliers. Because gap is conditioned on feasibility, we limit objective-quality claims and rely on feasibility and constraint-focused metrics for cross-variant analyses.}

\vspace{0.5em}
\noindent{\textbf{SCHEDBench-Variants:} To evaluate robustness across surface-form variants, we compare matched instances across different renderings. Since feasibility is binary and often zero on difficult families, we report two graded per-instance metrics: coverage and violation rate. For each model, instance, and variant, we extract three quantities: the number of assignments produced, the number required per instance, and number of violated hard constraints. Coverage is the proportion of required assignments produced; violation rate is the number of hard-constraint violations per produced assignment. As a single assignment may violate multiple constraints, violation rate is unbounded and interpreted only alongside coverage. To isolate individual surface-form axes, we use the matched 470-instance ablation set, varying one rendering axis while preserving the underlying formal instance. We pair outputs by source instance and compute per-instance changes in violation rate and coverage between the plain rendering v1 and each ablation variant. We summarize violation-rate shifts with the median and interquartile range of $|\Delta \mathrm{VR}|$, using absolute changes as improvements and degradations can cancel in signed averages. We also report $v_1$ violation rate as a capability anchor, so models with high violation rates are not treated as robust simply because their outputs vary little across renderings.  Per-instance changes are judged against the empirical seed band defined in Section~\ref{sec:seed}.}

\begin{table*}[t]
\centering
\small
\sisetup{table-number-alignment=center, detect-weight=true, detect-inline-weight=math}
\resizebox{\textwidth}{!}{%
\begin{tabular}{l S[table-format=3] S[table-format=1.3] l l l l c}
\toprule
\multicolumn{8}{c}{\textbf{v1 (plain) $\leftrightarrow$ v2 (constraint shuffle)}} \\
\midrule
\textbf{Model} & {$N^{\dagger}$} & {v1 viol.\ rate} & {$\bm{|\Delta\text{viol.\ rate}|}$} & {$\bm{|\Delta\text{coverage}|}$} & {Seed band $|\Delta\mathrm{vr}|$} & {95\% CI} & {Beyond noise?} \\
                & {\scriptsize paired} & {\scriptsize median} & {\scriptsize \textbf{median [Q1,\,Q3]}} & {\scriptsize \textbf{median [Q1,\,Q3]}} & {\scriptsize median [Q1,\,Q3]} & {\scriptsize $\Delta$med (eff$-$noise)} & \\
\midrule
meta-llama-4-maverick-17b-123e            & 470 & 0.272 & \textbf{0.116} {\scriptsize [0.032,\,0.254]} & 0.007 {\scriptsize [0.000,\,0.100]} & 0.075 {\scriptsize [0.013,\,0.213]} & [+0.018,\,+0.062] & \textbf{yes} \\
meta-llama-3.3-70b                        & 470 & 0.214 & \textbf{0.091} {\scriptsize [0.011,\,0.248]} & 0.017 {\scriptsize [0.000,\,0.100]} & 0.048 {\scriptsize [0.006,\,0.220]} & [+0.007,\,+0.073] & \textbf{yes} \\
claude-haiku-4-5 (2025-10-01)             & 470 & 0.315 & \textbf{0.072} {\scriptsize [0.011,\,0.187]} & 0.000 {\scriptsize [0.000,\,0.000]} & 0.045 {\scriptsize [0.006,\,0.139]} & [+0.001,\,+0.048] & \textbf{yes} \\
gemini-3.1-flash-lite                     & 468 & 0.143 & \textbf{0.062} {\scriptsize [0.011,\,0.170]} & 0.000 {\scriptsize [0.000,\,0.051]} & 0.034 {\scriptsize [0.001,\,0.156]} & [+0.010,\,+0.049] & \textbf{yes} \\
gpt-5.4-mini (2026-03-17)                 & 447 & 0.240 & \textbf{0.057} {\scriptsize [0.007,\,0.228]} & 0.000 {\scriptsize [0.000,\,0.000]} & 0.047 {\scriptsize [0.009,\,0.271]} & [-0.024,\,+0.038] & near \\
claude-sonnet-4-6                         & 465 & 0.132 & \textbf{0.053} {\scriptsize [0.011,\,0.136]} & 0.006 {\scriptsize [0.000,\,0.147]} & 0.064 {\scriptsize [0.010,\,0.164]} & [-0.029,\,+0.008] & near \\
qwen/qwen3.5-122b (2026-02-24)            & 466 & 0.080 & \textbf{0.045} {\scriptsize [0.000,\,0.156]} & 0.000 {\scriptsize [0.000,\,0.000]} & 0.062 {\scriptsize [0.004,\,0.219]} & [-0.030,\,+0.011] & near \\
qwen/qwen3.5-27b (2026-02-24)             & 470 & 0.045 & \textbf{0.045} {\scriptsize [0.000,\,0.188]} & 0.000 {\scriptsize [0.000,\,0.000]} & 0.038 {\scriptsize [0.000,\,0.190]} & [-0.020,\,+0.031] & near \\
gpt-5.4 (2026-03-05)                      & 445 & 0.143 & \textbf{0.041} {\scriptsize [0.010,\,0.113]} & 0.000 {\scriptsize [0.000,\,0.000]} & 0.025 {\scriptsize [0.005,\,0.094]} & [+0.008,\,+0.026] & \textbf{yes} \\
claude-opus-4-6                           & 467 & 0.074 & \textbf{0.031} {\scriptsize [0.006,\,0.083]} & 0.000 {\scriptsize [0.000,\,0.006]} & 0.029 {\scriptsize [0.006,\,0.081]} & [-0.003,\,+0.010] & near \\
qwen/qwen3.5-397b (2026-02-16)            & 467 & 0.033 & \textbf{0.031} {\scriptsize [0.000,\,0.104]} & 0.000 {\scriptsize [0.000,\,0.000]} & 0.011 {\scriptsize [0.000,\,0.081]} & [+0.000,\,+0.020] & near \\
gemini-3-flash-preview                    & 469 & 0.000 & \textbf{0.006} {\scriptsize [0.000,\,0.049]} & 0.000 {\scriptsize [0.000,\,0.008]} & 0.009 {\scriptsize [0.000,\,0.062]} & [-0.014,\,+0.006] & near \\
gpt-5.5 (2026-04-23)                      & 470 & 0.000 & \textbf{0.000} {\scriptsize [0.000,\,0.000]} & 0.000 {\scriptsize [0.000,\,0.000]} & 0.000 {\scriptsize [0.000,\,0.000]} & [+0.000,\,+0.000] & no \\
\midrule
\itshape Median across models              & 468 & 0.132 & \itshape \textbf{0.045} & \itshape 0.000 & --- & --- & --- \\
\bottomrule
\end{tabular}%
}
\caption{Per-instance volatility under \textbf{v2 (constraint shuffle)} on the 470-instance ablation set. $N^{\dagger}$ counts paired instances excluding no-output drops. The $|\Delta\text{viol.\ rate}|$ column reports absolute per-instance change; v1 violation rate serves as a capability anchor, since small changes can also indicate models are ``stably bad.'' Verdicts: \emph{yes}: CI strictly $>0$. \emph{no ($<$ noise)}: CI strictly $<0$. \emph{near}: CI straddles 0 with lower bound above $-$IQR of the seed band. \emph{no}: effect and seed band both identically zero (no effect, no measurable noise).}
\label{tab:variation-v2}
\end{table*}

\section{Results and Key Findings}

Our results show that LLM scheduling performance is not fully invariant to semantically equivalent renderings of the same underlying instances under tested surface-form perturbations. Across the matched 470-instance ablation set, constraint reordering induces per-instance changes beyond the seed-noise floor for several models, whereas theme and paraphrase variation produce shifts largely within it. Not only does this make the benchmark harder to solve; it indicates constraint ordering can act as a latent control variable over the generated schedule, altering the composition of outcomes.

\subsection{Full-Variation Performance}

Under full variation, GPT-5.5 is the strongest model tested, achieving 55.9\% feasibility, followed by Gemini 3 Flash Preview at 29.5\%; all remaining models fall below 26\%, with a 15.2\% cross-model average, indicating both frontier and large open-weight models often fail to produce complete feasible schedules with all variation axes enabled. Performance varies sharply by family and difficulty: INRC Easy reaches 97\% average feasibility, while INRC Hard falls to 1.7\%, and RCPSP-SM Hard is nearly unsolved at 0.3\%. JSP shows especially strong separation, with GPT-5.5 remaining near-perfect across difficulty levels while other models solve few or no instances. Relative to the matched simpler 470-instance plain-rendering subset ($v_1$), full variation adds a substantial surface-form challenge, reducing average feasibility from 26.1\% to 21.5\%, with especially large drops for GPT-5.5 ($84.0\%\rightarrow61.5\%$) and Gemini 3 Flash Preview ($52.1\%\rightarrow37.2\%$).

\subsection{Constraint Reordering}

Among the isolated surface-form axes, constraint reordering ($v_2$) produces the strongest above-noise sensitivity. Comparing v1 to v2, the median absolute violation-rate change across models is 0.045, with five models exceeding the seed-sensitivity noise band: Llama 4 Maverick (0.116), Llama 3.3 (0.091), Claude Haiku (0.072), Gemini Flash Lite (0.062), and GPT-5.4 (0.041), all exceeding the seed-sensitivity noise band. This is notable because $v_2$ changes only constraint order, leaving entity names, paraphrase templates, numerical values, feasible regions, and optimal objectives unchanged. Movement under $v_2$ suggests some models are sensitive not only to formal problem content, but also to the sequence in which constraints are encountered, making ordering a latent control variable over the generated schedule.

\subsection{Thematic Reframing}

Thematic variation ($v_{3A}$) replaces canonical labels with domain-framed vocabulary while preserving the underlying formal instance. It produces a small but measurable median absolute violation-rate change of 0.031 across models, while signed median changes remain near zero, indicating that domain framing shifts individual instances in both directions without systematically improving or degrading aggregate performance. Seed-sensitivity analysis shows that re-rendering the same instances under different themes produces comparable variation, suggesting that $v_{3A}$ is a low-magnitude perturbation mostly comparable to seed-level rendering variation, rather than an artifact of a particular sampled theme.

\subsection{Lexical-Syntactic Paraphrasing}

This condition isolates semantics-preserving lexical and syntactic variation while retaining the schematic structure of the formal instance. It produces a median absolute violation-rate change of 0.031 across models, matching $v_{3A}$, but most model-axis pairs remain near the seed-sensitivity noise floor, with only Llama 4 Maverick classified as beyond noise. These results suggest that paraphrase alone is not the dominant source of instability in SCHEDBench: rewording can change individual outputs, but for most models its aggregate effect is comparable to alternate paraphrase-templates. This contrasts with constraint reordering, where identical formal constraints produce clearer above-noise movement when presented in a different sequence.

\subsection{Feasibility and Objective Quality}

SCHEDBench shows that feasibility and objective quality are separable dimensions of schedule generation. Among feasible outputs, mean optimality gaps frequently far exceed median gaps, GPT-5.5 for instance, has a 4.3\% median but 747.1\% mean gap under v1, indicating heavy-tailed quality distributions in which most feasible schedules are near-optimal but a few are extreme, with this pattern holding across several models/conditions.  We therefore treat schedule validity and objective quality as complementary metrics, distinguishing that producing a valid schedule does not imply producing a good one. We report both metrics, as the gap is conditioned on feasibility, inherently dependent on the instances a model solves.

\subsection{Main Findings}

The results support four main conclusions. First, LLM scheduling behavior is not fully stable under semantics-preserving rendering variation. On matched formal instances, re-rendering shifts per-instance violation rates above a seed-noise floor for a subset of models. This is the central empirical finding of SCHEDBench. Second, surface-form variation induces instance-level changes not captured by signed aggregate differences alone. Signed effects are near-zero, but absolute per-instance changes are often substantial, indicating models may improve on some instances while failing on others. Third, constraint reordering is the only tested surface-form axis that consistently produces above-noise sensitivity for a subset of models. Thematic and paraphrase variation induce visible per-instance movement, but their effects fall largely within the rendering-seed band. Finally, feasibility alone is insufficient to characterize model behavior. Domain verifiers expose violations and optimality gaps, allowing SCHEDBench to distinguish invalid schedules, feasible but low-quality schedules, and schedules whose constraint-violation profile changes under equivalent renderings. Overall, SCHEDBench shows that current LLMs are not reliably invariant to natural-language presentations of the same combinatorial scheduling problem. The strongest evidence is not merely that performance decreases under full variation, but that surface-form changes reshape constraint satisfaction behavior at the instance level.

\section{Conclusion}

We introduce SCHEDBench, a natural-language benchmark for combinatorial scheduling built from canonical scheduling instances. SCHEDBench spans six problem families, paired with solver-verified feasibility and optimality references. Evaluating thirteen frontier and open-weight LLMs, we find that current models largely fail this task, and do not maintain fully invariant constraint-satisfaction behavior under surface-form variation.

\section*{Limitations}

SCHEDBench evaluates direct schedule generation; invariance claims are limited to final behavior and do not isolate effects to the modeling stage. Limited exploratory experiments prompting models to emit solver-code are additionally reported in the appendix; broader investigation in this setting and other generation paradigms (including evaluation of intermediate formal representations) are left to future work. The benchmark is also built using a template-based generation pipeline. While this provides strong experimental control, it also produces more regular language. Expanding SCHEDBench with broader human-authored phrasing and other natural-language renderings may further improve its ecological validity. We evaluate models under default inference with greedy decoding, and extended-reasoning modes may further improve rendering invariance, though not a central limitation; as reported sensitivity is already weakest among the strongest models. Whether extended reasoning attenuates it is left to future work.

\bibliography{main}

\appendix

\section{Benchmark Construction Details}
\label{sec:appendixA}

We provide the code for all parts of the benchmark generation pipeline at \url{https://github.com/SCHEDBench/SCHEDBench_Datagen}, and the full 1,132 instance SCHEDBench evaluation set used in Table \ref{tab:v4-difficulty}  at \url{https://huggingface.co/datasets/SCHEDBench/SCHEDBench}.

\subsection{Instance Families and Selection Criteria}
\label{sec:appendixA-instance-families}

Table~\ref{tab:appendixA-instance-families} summarizes the source, count, selection methodology, and size range for each family.

\begin{table*}[t]
\centering
\small
\setlength{\tabcolsep}{6pt}
\begin{tabularx}{\textwidth}{@{}l l c l X@{}}
\toprule
\textbf{Problem Type} & \textbf{Source Repository} & \textbf{Count} & \textbf{Selection Strategy} & \textbf{Structural Dimensions / Size Range} \\
\midrule
JSP            & JSPLib           & 262 & Canonical Subset  & $6\times6$ to $100\times100$ (Jobs $\times$ Machines) \\
RCPSP-SM        & PSPLib           & 400 & Alphabetic Subset & 32--122 Activities, 4 Renewable Resources \\
RCPSP-MM        & PSPLib           & 300 & Alphabetic Subset & 12--32 Activities, Mixed Multi-Mode Resources \\
Timetabling     & ITC-2007 Track 3 &  21 & Exhaustive Census & 30--131 Courses, 5--20 Rooms \\
Nurse Rostering & INRC-I           &  69 & Exhaustive Census & 10--50 Employees, 28-day Horizon \\
Nurse Rostering & NSPLib (N25)     &  80 & Alphabetic Subset & 25 Nurses, 4 Shifts, 7-day Horizon \\
\bottomrule
\end{tabularx}
\caption{Canonical scheduling source families, sampling criteria, and instance dimensions in \textsc{SCHEDBench}.}
\label{tab:appendixA-instance-families}
\end{table*}

\paragraph{JSPLib}
We sample across nine JSPLib sub-families, with (N) instances provided from each: 
\texttt{abz}~(5)~\cite{adams1988shifting}, 
\texttt{dmu}~(80)~\cite{demirkol1998benchmarks}, 
\texttt{ft}~(3)~\cite{fisher1963probabilistic}, 
\texttt{la}~(40)~\cite{lawrence1984resource}, 
\texttt{orb}~(10)~\cite{applegate1991computational}, 
\texttt{swv}~(20)~\cite{storer1992new}, 
\texttt{ta}~(70)~\cite{taillard1993benchmarks}, 
\texttt{tai}~(30)~\cite{dacol2022industrial}, and 
\texttt{yn}~(4)~\cite{yamada1992genetic}.
A fully compiled JSPLib for instances and BKS was accessed from: \url{https://scheduleopt.github.io/benchmarks/jsplib/#jobshop-benchmark-instances}. 

\paragraph{RCPSP-SM}
We take the first 100 instances per size class (\texttt{j30}, \texttt{j60}, \texttt{j90}, \texttt{j120}) via alphanumeric filename ordering.
RCPSP-SM Instances and BKS were accessed from: \url{https://www.om-db.wi.tum.de/psplib/data.php}

\paragraph{RCPSP-MM}
The same alphanumeric selection rule is applied across three size classes (\texttt{j10}, \texttt{j20}, \texttt{j30}), totaling 300 instances.
RCPSP-MM Instances and BKS were accessed from: \url{https://www.om-db.wi.tum.de/psplib/data.php}

\paragraph{ITC}
All 21 instances (\texttt{comp01}--\texttt{comp21}) from ITC-2007 Track 3 are included exhaustively.
ITC Instances and BKS were accessed from: \url{https://github.com/Docheinstein/itc2007-cct/tree/master/datasets} and remaining solutions from \cite{hao_benlic_2011_itc2007_lower_bounds}.

\paragraph{INRC}
The complete 69-instance INRC-I corpus partitions into 33 sprint, 18 medium, and 18 long sub-problems. INRC Instances and BKS were accessed from: \url{https://benchmark.gent.cs.kuleuven.be/nrp/}.

\paragraph{NSPLib}
We cross-reference instances \texttt{n25\_1}--\texttt{n25\_10} against evaluation scenarios 1--8, yielding 80 instances with fixed structure (25 nurses, 4 shifts, 7-day horizon) and varying constraint densities.
NSPLib Instances and BKS were accessed from: \url{https://www.projectmanagement.ugent.be/research/personnel_scheduling/nsp}.

\subsection{Difficulty Selection}
\label{sec:appendixA-difficulty}

Difficulty tiers (\textit{Easy}, \textit{Medium}, \textit{Hard}) are assigned independently within each family and reflect intra-domain complexity; they are not cross-comparable across families.

\paragraph{JSPLib}
Difficulty is inherited from JSPLib meta-documentation, based on empirical solver hardness and the proximity of best-known upper bounds to proven lower bounds.

\paragraph{RCPSP (Single and Multi-Mode)}
Difficulty maps directly to instance size class. For RCPSP-SM: easy (\texttt{j30}), medium (\texttt{j60}), hard (\texttt{j90} and \texttt{j120}); for RCPSP-MM: easy (\texttt{j10}), medium (\texttt{j20}), hard (\texttt{j30}).

\paragraph{INRC}
Difficulty is inherited from the canonical source: sprint (easy), medium, and long (hard).

\paragraph{ITC and NSPLib}
Lacking standardized size partitions, difficulty tiers are derived by ranking all instances within each family by their canonical BKS objective value and segmenting into three equal-sized percentile bins.

\subsection{Verbalization Templates}
\label{sec:appendixA-verbalization-templates}

A seeded pseudorandom function selects one template variant per constraint entry, independently of entity mapping and constraint ordering. Table~\ref{tab:appendix-template-exemplars} shows three representative exemplars per family across all constraint categories; Table~\ref{tab:appendixA-template-bank-summary} summarizes total template coverage by domain.

\begin{table}[htbp]
\centering
\small
\begin{tabular*}{\columnwidth}{@{\extracolsep{\fill}} l c c @{}}
\toprule
\textbf{Family} &
\textbf{\shortstack{Constraint\\Categories}} &
\textbf{\shortstack{Templates per\\Category}} \\
\midrule
JSPLib   &  3 & 6 \\
RCPSP-SM &  8 & 5--21 \\
RCPSP-MM & 10 & 5--12 \\
ITC      &  7 & 4--8 \\
INRC     & 21 & 4--7 \\
NSPLib   & 15 & 4--5 \\
\bottomrule
\end{tabular*}
\caption{Template bank coverage by domain (including category subtypes).}
\label{tab:appendixA-template-bank-summary}
\vspace{-0.5em}
\end{table}

\begin{table*}[!t]
\centering
\small
\setlength{\tabcolsep}{5pt}
\renewcommand{\arraystretch}{1.3}
\begin{tabularx}{\textwidth}{@{} l l X @{}}
\toprule
\textbf{Family} & \textbf{Constraint Type} & \textbf{Template Example} \\
\midrule
\textbf{JSPLib}     & Per-Operation (Var.\,A) & \texttt{\{job\_name\} operation \{op\_id\} must run on \{machine\_name\} for \{duration\_text\}.} \\
                  & Per-Operation (Var.\,B) & \texttt{For \{item\_label\} \{job\_id\}, operation \{op\_id\} occupies \{location\_label\} \{machine\_id\} for \{duration\_int\} time units.} \\
                  & Per-Operation (Var.\,C) & \texttt{Operation \{op\_id\} of job \{job\_id\} requires machine \{machine\_id\} for \{duration\_int\} time units.} \\
\addlinespace[4pt]
\textbf{RCPSP-SM} & Resource Capacity   & \texttt{\{resource\} has an operational capacity of \{capacity\_int\} units.} \\
                  & Activity Profile    & \texttt{\{activity\} requires a span of \{duration\_phrase\} and consumes \{demand\_text\} uniformly.} \\
                  & Precedence Relation & \texttt{\{predecessor\} must completely finish before \{successor\} is allowed to commence.} \\
\addlinespace[4pt]
\textbf{RCPSP-MM} & Mode Selection      & \texttt{Activity \{activity\} must be locked into exactly one processing mode.} \\
                  & Mode Definition     & \texttt{Under mode \{mode\}, \{activity\} runs for \{duration\_phrase\}, requiring \{renewable\_clause\}...} \\
                  & Precedence Relation & \texttt{\{successor\} is restricted from starting until \{predecessor\} has concluded.} \\
\addlinespace[4pt]
\textbf{ITC}      & Course Specification & \texttt{\{activity\} is taught by \{worker\}, spans \{sessions\} sessions, and caps at \{capacity\}.} \\
                  & Curriculum Conflict  & \texttt{Curriculum \{crew\} links \{course\_list\}; no elements may overlap temporally.} \\
                  & Availability         & \texttt{\{activity\} is unavailable for scheduling on Day \{day\}, Period \{period\}.} \\
\addlinespace[4pt]
\textbf{INRC}     & Contractual Policy  & \texttt{No \{worker\_label\} may exceed a working block of \{value\} consecutive days.} \\
                  & Demand Coverage     & \texttt{For each \{day\_of\_week\}, a minimum of \{worker\_count\} personnel must be active.} \\
                  & Staffing Request    & \texttt{\{worker\_name\} has submitted a preference against assignment to \{shift\_ref\}...} \\
\addlinespace[4pt]
\textbf{NSPLib}   & Target Coverage     & \texttt{On day \{day\}, exactly \{required\_count\} staff member(s) must cover the shift.} \\
                  & Preference Costing  & \texttt{Assigning \{worker\_name\} to \{shift\}... incurs a preference cost of \{cost\}.} \\
                  & Consecutive Limits  & \texttt{Any continuous working stretch for a \{worker\_label\} must fall within a window...} \\
\bottomrule
\end{tabularx}

\caption{Representative template exemplars across all six benchmark families and constraint types.}
\label{tab:appendix-template-exemplars}
\end{table*}

\subsection{Entity Lexicons and Domain Lists}
\label{sec:appendixA-entity-lexicons}

Each instance is mapped to a distinct semantic theme via a seeded hash function, assigned independently of template selection and constraint ordering. For larger RCPSP instances, entity names are augmented with scale modifiers; INRC shift labels are deterministically mapped to thematic aliases; NSPLib uses canonical shift markers with domain-specific off-duty variants. Table~\ref{tab:appendixA-themes-entity-pools} summarizes theme assignments and pool sizes.

\subsection{Controlled Variation Axes}
\label{sec:appendixA-variation-axes}

Each generation path is bound to an isolated random seed (seed 42 is the benchmark standard); all three axes are independently disableable to produce the ablation conditions in Section~4. The following examples use \texttt{ft06} as a reference instance.

\paragraph{Template Realization}
\begin{quote}
\small
\textbf{Seed A:} \textit{``The manufacturing layout specifies that 6 batches must be processed across 6 distinct workstations...''} \\
\textbf{Seed B:} \textit{``An analysis of the manufacturing floor indicates a workload of 6 batches requiring allocation on 6 workstations...''}
\end{quote}

\paragraph{Thematic Domain}
\begin{quote}
\small
\textbf{Manufacturing:} \textit{``Batch Trenton must occupy the South Paint Booth for a period of 3 time units...''} \\
\textbf{Logistics:} \textit{``Shipment Bayside must occupy the Local Freight Lane for a period of 3 time units...''}
\end{quote}

\paragraph{Constraint Sequence}
\begin{quote}
\small
\textbf{Permutation X:} \textit{[Batch Trenton statement] precedes [Batch Sunnyvale statement]...} \\
\textbf{Permutation Y:} \textit{[Batch Sunnyvale statement] precedes [Batch Trenton statement]...}
\end{quote}
For \texttt{ft06}, job-order permutations yield $6! = 720$ distinct prompt orderings.

\subsection{Faithfulness Audit}
\label{sec:appendixA-faithfulness-audit}

Two expert annotators independently verified a stratified sample of 100 instances across all six families and three difficulty tiers, checking each verbalized prompt against its canonical source record for numerical durations, precedence relations, resource constraints, and problem conditions. Discrepancies were resolved through joint reconciliation. The audited sample achieved zero semantic errors, confirming pipeline fidelity.



\subsection{BKS Source Documentation}
\label{sec:appendixA-bks-source-documentation}

We re-ran all instances with OR-Tools CP-SAT to verify reference BKS bounds; no local solve found an objective value better than the canonical BKS at the time of writing. Automated validation uses OR-Tools CP-SAT (v9.15.6755) across 8 parallel workers, with family-specific wall-clock timeouts: 10 seconds for JSPLib, RCPSP-SM, RCPSP-MM, and NSPLib; 20 seconds for ITC and INRC. Instances where the solver does not return a feasibility verdict within the timeout retain their BKS value and remain in the evaluation set.

\subsection{Artifact Licenses}
\label{sec:appendixA-license-documentation}
Several canonical benchmark suites used to construct
SCHEDBench (including PSPLib, JSPLib, ITC-2007,
INRC-I, and NSPLib) do not specify explicit modern
redistribution licenses. This is common for historical
operations-research benchmarks released for open
academic research. To respect original provenance,
we do not directly redistribute the canonical source
matrices; they can instead be obtained from the
public repositories listed earlier in Appendix A. 

SCHEDBench preserves the original research-oriented
usage context of these benchmark suites. Released
artifacts consist only of natural-language derivative
renderings for language-model evaluation and
reproducibility research. The generation pipeline is released under the MIT License, while the derived benchmark is released under CC BY 4.0 to support transparent reuse and reproducibility. To the best of our knowledge, these releases and their intended research use are compatible with the publicly accessible research-use conditions associated with the original benchmark sources.

\section{Model and Inference Configuration}
\label{sec:appendixB}
\subsection{Reproducibility}

\paragraph{Models and Compute (C1)} All models were accessed via
provider hosted inference APIs. Inference was parallelized on CPU-only machines, with each model evaluated by six worker processes each issuing up to nine concurrent API requests. Evaluated models span OpenAI, Anthropic, Google, Qwen~3.5, and Meta
families; open-weight sizes are
as named, proprietary sizes undisclosed. Evaluation consumed approximately
490M input and 920M output tokens (approx. 1.4B total). Based on publicly available on-demand API pricing as of April 23, 2026, and aggregate token volume reported above, the total inference cost of evaluating SCHEDBench is estimated at approximately \$2,371 : \$1,271 for OpenAI models, \$971 for Anthropic models, \$64 for Qwen models, \$36 for Google Gemini models, and \$29 for Meta Llama models as tested; noting that open-source model costs vary by deployment infrastructure and associated operating costs.

\paragraph{Setup (C2)} All models were evaluated zero-shot with
greedy decoding (temperature~0) under default settings. No hyperparameter
search or model selection was performed; median per-query context is
approximately 13.8k tokens.

\paragraph{Reporting (C3)} Each model--instance pair is evaluated
once; greedy decoding renders runs deterministic. Variability is assessed
via the seed-sensitivity ablation (Section~\ref{sec:seed}) and bootstrap
95\% confidence intervals.

\paragraph{Software (C4)} Feasibility is verified using OR-Tools
CP-SAT (v9.15.6755); reference objectives are best-known solutions from the
source benchmark libraries. Code and data are released.

\paragraph{Inference APIs}

\begin{itemize}
    \item \textbf{Claude:} \href{https://platform.claude.com/docs/en/home}{Claude API}
    \item \textbf{GPT:} \href{https://developers.openai.com/api/docs}{OpenAI API}
    \item \textbf{Gemini:} \href{https://ai.google.dev/gemini-api/docs}{Gemini API}
    \item \textbf{Llama:} \href{https://docs.aws.amazon.com/bedrock/latest/APIReference/API_Operations_Amazon_Bedrock_Runtime.html}{Amazon Bedrock Runtime API}
    \item \textbf{Qwen:} \href{https://www.alibabacloud.com/help/en/model-studio/first-api-call-to-qwen}{Alibaba Cloud Model Studio / DashScope API}
\end{itemize}

\subsection{System Prompts}

A single system prompt is applied to every model and instance:

\begin{quote}\small\ttfamily
You are completing an automated benchmark. The user message contains a scheduling problem and an exact output format. Begin your response with the first schedule line and output only the schedule in that format --- nothing else: no preamble, no explanation, no reasoning, no markdown formatting (no \textbf{bold}, no \texttt{inline code}, no triple-backtick code blocks), no XML tags such as \texttt{<schedule>} or \texttt{<answer>}, no tool tags, no Python code. Do not call any tools, do not invoke external solvers, and do not generate code to be executed --- solve the problem yourself using only your own reasoning. Your response is fed directly into a parser; any extra characters cause the response to be discarded.
\end{quote}

\noindent{The user message appends a family-specific \emph{Response Format} stanza fixing the per-domain output grammar. Format heads for each of the six families are shown below; complete rule lists are in the released code (\texttt{src/domains/<family>/assets/
prompt\_text.txt})}

\paragraph{JSPLib (JSP).}
\begin{verbatim}
Response Format:
<ItemName> step 1: start=<integer>
<ItemName> step 2: start=<integer>
...
Example: "Shipment Madison step 1: start=0"
\end{verbatim}

\paragraph{RCPSP-SM.}
\begin{verbatim}
Response Format:
<ActivityName>: <integer>
Example: "Project Kickoff: 0"
\end{verbatim}

\paragraph{RCPSP-MM.}
\begin{verbatim}
Response Format:
<ActivityName>: start=<integer>, mode=<integer>
Example: "Methods Audit: start=11, mode=1"
\end{verbatim}

\paragraph{NSPLib.}
\begin{verbatim}
Response Format:
<WorkerIdentifier>|<Day>|<ShiftId>
Example: "Ari|1|service slot 1"
\end{verbatim}

\paragraph{INRC.}
\begin{verbatim}
Response Format:
ASSIGNMENTS=<count>
<DAY_OF_MONTH>,<EmployeeName>,
<ShiftTypeName>;
Example: "1,Teresa,Control Room Day;"
\end{verbatim}

\paragraph{ITC.}
\begin{verbatim}
Response Format:
<ActivityName>|<LocationName>|<Day>|
<Day_Period>
Example: "Mathematics 101|Room A|0|3"
\end{verbatim}

\subsection{Model Identifiers}

API model strings used in every request, grouped by provider. 

\paragraph{OpenAI.}
\texttt{gpt-5.4-mini-2026-03-17},
\texttt{gpt-5.4-2026-03-05},
\texttt{gpt-5.5-2026-04-23}.

\paragraph{Anthropic.}
\texttt{claude-haiku-4-5-20251001},
\texttt{claude-sonnet-4-6},
\texttt{claude-opus-4-6}.

\paragraph{Google Gemini.}
\texttt{gemini-3-flash-preview},
\texttt{gemini-3.1-flash-lite}.

\paragraph{Qwen.}
\texttt{qwen/qwen3.5-397b-a17b-20260216},
\texttt{qwen/qwen3.5-122b-a10b-20260224},
\texttt{qwen/qwen3.5-27b-20260224}.

\paragraph{Meta Llama.}
\texttt{meta.llama4-maverick-17b-
instruct-v1:0},
\texttt{meta.llama3-3-70b-instruct
-v1:0}.

\vspace{.5em}  
\noindent\footnotesize\textit{Note on Anthropic IDs.} For the 4.6 generation, the dateless IDs (\texttt{claude-sonnet-4-6}, \texttt{claude-opus-4-6}) are pinned snapshots mapping to a single fixed set of weights; they are not evergreen aliases. New versions ship under new IDs.



\begin{table*}[!t]
\centering
\small
\setlength{\tabcolsep}{5pt}
\begin{tabularx}{\textwidth}{@{} l l c c @{}}
\toprule
\textbf{Family} & \textbf{Assigned Thematic Domains} & \textbf{Job/Task Pool} & \textbf{Asset/Resource Pool} \\
\midrule
JSPLib     & Manufacturing, Logistics, Film, etc.\ (10 domains)     & 100 names       & 100 per theme \\
RCPSP-SM & Workflow, Software Dev, Construction, Research          & 60 names        & Modifier-augmented \\
RCPSP-MM & Workflow, Software Dev, Construction, Research          & 36 names        & Segmented pools \\
ITC      & Film Dev, Game Studio, Consulting, Lab Ops, Flight      & Per-theme tasks & Per-theme rooms \\
INRC     & Corporate Security, Commercial Retail                   & 60 names        & Shift aliases \\
NSPLib   & Warehouse, Hospitality, Clinic, Airport Ground, Retail  & 40 names        & Static labels \\
\bottomrule
\end{tabularx}
\caption{Thematic domain assignments and entity lexicon pool sizes across problem families.}
\label{tab:appendixA-themes-entity-pools}
\end{table*}

\begin{table*}[t]
\centering
\small
\sisetup{table-number-alignment=center, detect-weight=true, detect-inline-weight=math}
\resizebox{\textwidth}{!}{%
\begin{tabular}{l S[table-format=3.1] S[table-format=3.1] S[table-format=3.1] S[table-format=3.1] S[table-format=3.1] S[table-format=3.1] S[table-format=3.1] S[table-format=4.1] S[table-format=3.1]}
\toprule
& {\textbf{Overall}} & \multicolumn{6}{c}{\textbf{Feasibility by scheduling family} (\%)} & \multicolumn{2}{c}{\textbf{Gap on feasible} (\%)} \\
\cmidrule(lr){3-8} \cmidrule(lr){9-10}
\textbf{Model} & {\textbf{Feas.\ \%}} & {ITC} & {INRC} & {NSPLib} & {JSPLib} & {RCPSP-MM} & {RCPSP-SM} & {Mean} & {Median} \\
                & & {($n_d{=}21$)} & {($n_d{=}69$)} & {($n_d{=}80$)} & {($n_d{=}100$)} & {($n_d{=}100$)} & {($n_d{=}100$)} & & \\
\midrule
gpt-5.5 (2026-04-23)                      & \bfseries 84.0 & \bfseries 4.8 & \bfseries 65.2 & \bfseries 100.0 & \bfseries 98.0 & 89.0 & \bfseries 82.0 & 747.1 & \bfseries 4.3 \\
gemini-3-flash-preview                    & 52.1 & 0.0 & \bfseries 65.2 & 85.0 & 11.0 & \bfseries 91.0 & 30.0 & \bfseries 95.1 & 18.2 \\
qwen/qwen3.5-27b (2026-02-24)             & 39.8 & 0.0 & \bfseries 65.2 & 77.5 & 42.0 & 29.0 & 9.0 & 311.2 & 12.5 \\
qwen/qwen3.5-397b (2026-02-16)            & 38.7 & 0.0 & \bfseries 65.2 & 72.5 & 30.0 & 39.0 & 10.0 & 243.2 & 8.3 \\
qwen/qwen3.5-122b (2026-02-24)            & 28.9 & 0.0 & 62.3 & 65.0 & 3.0 & 30.0 & 8.0 & 254.4 & 8.7 \\
claude-opus-4-6                           & 18.3 & 0.0 & 62.3 & 10.0 & 0.0 & 24.0 & 11.0 & 197.6 & 56.2 \\
claude-sonnet-4-6                         & 18.1 & 0.0 & \bfseries 65.2 & 30.0 & 0.0 & 15.0 & 1.0 & 600.5 & 100.0 \\
gpt-5.4 (2026-03-05)                      & 11.5 & 0.0 & 55.1 & 0.0 & 0.0 & 9.0 & 7.0 & 893.1 & 227.5 \\
gemini-3.1-flash-lite                     & 10.2 & 0.0 & \bfseries 65.2 & 0.0 & 0.0 & 1.0 & 2.0 & 984.6 & 539.1 \\
meta-llama-3.3-70b                        & 10.0 & 0.0 & \bfseries 65.2 & 0.0 & 0.0 & 2.0 & 0.0 & 1500.8 & 979.7 \\
gpt-5.4-mini (2026-03-17)                 & 9.8 & 0.0 & \bfseries 65.2 & 0.0 & 0.0 & 1.0 & 0.0 & 1317.3 & 594.0 \\
claude-haiku-4-5 (2025-10-01)             & 9.6 & 0.0 & \bfseries 65.2 & 0.0 & 0.0 & 0.0 & 0.0 & 1550.4 & 616.3 \\
meta-llama-4-maverick-17b-123e            & 8.5 & 0.0 & 58.0 & 0.0 & 0.0 & 0.0 & 0.0 & 969.9 & 437.5 \\
\midrule
\itshape Average across models   & \itshape 26.1 & \itshape 0.4 & \itshape 63.4 & \itshape 33.8 & \itshape 14.2 & \itshape 25.4 & \itshape 12.3 & \itshape 743.5 & \itshape 277.1 \\
\bottomrule
\end{tabular}%
}
\caption{Per-model results on the \textbf{v1 (plain)} variant of \textsc{SchedBench} (the plain baseline rendering, with canonical entity names and no constraint reordering). Feasibility rate is shown overall and per scheduling family; mean and median percent optimality gap are computed over feasible outputs only ($(z - z^{\star})/z^{\star} \times 100$). Higher feasibility is better; lower gap is better. The best value in each column is shown in \textbf{bold}. Per-family instance counts $n_d$ are listed below the column headers; $n=470$ total.}
\label{tab:v1-per-model}
\end{table*}

\begin{table*}[t]
\centering
\small
\sisetup{table-number-alignment=center, detect-weight=true, detect-inline-weight=math}
\resizebox{\textwidth}{!}{%
\begin{tabular}{l S[table-format=3.1] S[table-format=3.1] S[table-format=3.1] S[table-format=3.1] S[table-format=3.1] S[table-format=3.1] S[table-format=3.1] S[table-format=4.1] S[table-format=3.1]}
\toprule
& {\textbf{Overall}} & \multicolumn{6}{c}{\textbf{Feasibility by scheduling family} (\%)} & \multicolumn{2}{c}{\textbf{Gap on feasible} (\%)} \\
\cmidrule(lr){3-8} \cmidrule(lr){9-10}
\textbf{Model} & {\textbf{Feas.\ \%}} & {ITC} & {INRC} & {NSPLib} & {JSPLib} & {RCPSP-MM} & {RCPSP-SM} & {Mean} & {Median} \\
                & & {($n_d{=}21$)} & {($n_d{=}69$)} & {($n_d{=}80$)} & {($n_d{=}262$)} & {($n_d{=}300$)} & {($n_d{=}400$)} & & \\
\midrule
gpt-5.5 (2026-04-23)                      & \bfseries 55.9 & 4.8 & 63.8 & \bfseries 91.2 & \bfseries 98.1 & \bfseries 56.7 & \bfseries 22.0 & 740.6 & 10.5 \\
gemini-3-flash-preview                    & 29.5 & 0.0 & 59.4 & 85.0 & 9.5 & 49.7 & 12.8 & \bfseries 118.3 & 19.0 \\
qwen/qwen3.5-27b (2026-02-24)             & 25.0 & 0.0 & 63.8 & 60.0 & 45.8 & 21.3 & 1.8 & 444.5 & 216.0 \\
qwen/qwen3.5-397b (2026-02-16)            & 24.3 & 0.0 & 65.2 & 60.0 & 26.7 & 30.3 & 5.2 & 288.6 & 8.0 \\
qwen/qwen3.5-122b (2026-02-24)            & 14.3 & 0.0 & 58.0 & 62.5 & 3.8 & 18.3 & 1.8 & 280.2 & \bfseries 7.5 \\
claude-sonnet-4-6                         & 12.8 & \bfseries 9.5 & 65.2 & 17.5 & 4.2 & 23.7 & 0.5 & 448.1 & 36.7 \\
claude-opus-4-6                           & 8.9 & 0.0 & 65.2 & 3.8 & 0.0 & 15.0 & 2.0 & 351.9 & 92.3 \\
claude-haiku-4-5 (2025-10-01)             & 4.9 & 0.0 & \bfseries 69.6 & 0.0 & 0.0 & 2.7 & 0.0 & 1187.0 & 504.3 \\
meta-llama-3.3-70b                        & 4.9 & 0.0 & 63.8 & 0.0 & 0.0 & 4.0 & 0.0 & 1202.0 & 389.3 \\
meta-llama-4-maverick-17b-123e            & 4.5 & 0.0 & 63.8 & 0.0 & 0.0 & 2.3 & 0.0 & 1207.8 & 600.0 \\
gpt-5.4-mini (2026-03-17)                 & 4.3 & 0.0 & 66.7 & 0.0 & 0.0 & 1.0 & 0.0 & 1371.7 & 634.0 \\
gemini-3.1-flash-lite                     & 3.9 & 0.0 & 59.4 & 0.0 & 0.0 & 1.0 & 0.0 & 1225.4 & 558.1 \\
gpt-5.4 (2026-03-05)                      & 3.7 & 0.0 & 56.5 & 0.0 & 0.0 & 1.0 & 0.0 & 1234.7 & 397.7 \\
\midrule
\itshape Average across models   & \itshape 15.2 & \itshape 1.1 & \itshape 63.1 & \itshape 29.2 & \itshape 14.5 & \itshape 17.5 & \itshape 3.5 & \itshape 777.0 & \itshape 267.2 \\
\bottomrule
\end{tabular}%
}
\caption{Per-model results on the \textbf{v4 (Full-variation)} variant of \textsc{SchedBench} (all three surface-form axes (themed entity names, paraphrased templates, and constraint reordering) active simultaneously). Feasibility rate is shown overall and per scheduling family; mean and median percent optimality gap are computed over feasible outputs only ($(z - z^{\star})/z^{\star} \times 100$). Higher feasibility is better; lower gap is better. The best value in each column is shown in \textbf{bold}. Per-family instance counts $n_d$ are listed below the column headers; $n=1132$ total.}
\label{tab:v4-per-model}
\end{table*}

\begin{table*}[t]
\centering
\small
\sisetup{table-number-alignment=center, detect-weight=true, detect-inline-weight=math}
\resizebox{\textwidth}{!}{%
\begin{tabular}{l S[table-format=3.1] S[table-format=3.1] S[table-format=3.1] S[table-format=3.1] S[table-format=3.1] S[table-format=3.1] S[table-format=3.1] S[table-format=4.1] S[table-format=3.1]}
\toprule
& {\textbf{Overall}} & \multicolumn{6}{c}{\textbf{Feasibility by scheduling family} (\%)} & \multicolumn{2}{c}{\textbf{Gap on feasible} (\%)} \\
\cmidrule(lr){3-8} \cmidrule(lr){9-10}
\textbf{Model} & {\textbf{Feas.\ \%}} & {ITC} & {INRC} & {NSPLib} & {JSPLib} & {RCPSP-MM} & {RCPSP-SM} & {Mean} & {Median} \\
                & & {($n_d{=}21$)} & {($n_d{=}69$)} & {($n_d{=}80$)} & {($n_d{=}100$)} & {($n_d{=}100$)} & {($n_d{=}100$)} & & \\
\midrule
gpt-5.5 (2026-04-23)                      & \bfseries 61.5 & 4.8 & 63.8 & \bfseries 91.2 & \bfseries 98.0 & \bfseries 53.0 & \bfseries 20.0 & 466.8 & 9.5 \\
gemini-3-flash-preview                    & 37.2 & 0.0 & 59.4 & 85.0 & 11.0 & 43.0 & 12.0 & \bfseries 183.0 & 12.8 \\
qwen/qwen3.5-27b (2026-02-24)             & 34.7 & 0.0 & 63.8 & 60.0 & 52.0 & 15.0 & 4.0 & 498.1 & 215.7 \\
qwen/qwen3.5-397b (2026-02-16)            & 32.1 & 0.0 & 65.2 & 60.0 & 24.0 & 26.0 & 8.0 & 384.2 & 14.1 \\
qwen/qwen3.5-122b (2026-02-24)            & 23.6 & 0.0 & 58.0 & 62.5 & 2.0 & 17.0 & 2.0 & 382.4 & \bfseries 8.8 \\
claude-sonnet-4-6                         & 18.3 & \bfseries 9.5 & 65.2 & 17.5 & 6.0 & 18.0 & 1.0 & 737.0 & 355.4 \\
claude-opus-4-6                           & 13.2 & 0.0 & 65.2 & 3.8 & 0.0 & 11.0 & 3.0 & 539.0 & 242.5 \\
claude-haiku-4-5 (2025-10-01)             & 10.9 & 0.0 & \bfseries 69.6 & 0.0 & 0.0 & 3.0 & 0.0 & 1293.3 & 516.9 \\
gpt-5.4-mini (2026-03-17)                 & 10.2 & 0.0 & 66.7 & 0.0 & 0.0 & 2.0 & 0.0 & 1396.7 & 652.5 \\
meta-llama-3.3-70b                        & 10.2 & 0.0 & 63.8 & 0.0 & 0.0 & 4.0 & 0.0 & 1375.1 & 525.1 \\
meta-llama-4-maverick-17b-123e            & 9.6 & 0.0 & 63.8 & 0.0 & 0.0 & 1.0 & 0.0 & 1355.6 & 676.4 \\
gemini-3.1-flash-lite                     & 8.9 & 0.0 & 59.4 & 0.0 & 0.0 & 1.0 & 0.0 & 1279.6 & 564.5 \\
gpt-5.4 (2026-03-05)                      & 8.5 & 0.0 & 56.5 & 0.0 & 0.0 & 1.0 & 0.0 & 1294.6 & 433.5 \\
\midrule
\itshape Average across models   & \itshape 21.5 & \itshape 1.1 & \itshape 63.1 & \itshape 29.2 & \itshape 14.8 & \itshape 15.0 & \itshape 3.8 & \itshape 860.4 & \itshape 325.2 \\
\bottomrule
\end{tabular}%
}
\caption{Per-model results on the \textbf{v4 (Full-variation), on v1's 470-instance subset} variant of \textsc{SchedBench} (all three surface-form axes active, restricted to the same 470 source\_instances as v1 so every cell is directly comparable to Table~\ref{tab:v1-per-model} model-by-model). Feasibility rate is shown overall and per scheduling family; mean and median percent optimality gap are computed over feasible outputs only ($(z - z^{\star})/z^{\star} \times 100$). Higher feasibility is better; lower gap is better. The best value in each column is shown in \textbf{bold}. Per-family instance counts $n_d$ are listed below the column headers; $n=470$ total.}
\label{tab:v4-per-model-on-v1-subset}
\end{table*}

\begin{table*}[t]
\centering
\small
\sisetup{table-number-alignment=center, detect-weight=true, detect-inline-weight=math}
\resizebox{\textwidth}{!}{%
\begin{tabular}{l S[table-format=3.1] S[table-format=3.1] S[table-format=3.1] S[table-format=3.1] S[table-format=3.1] S[table-format=3.1] S[table-format=3.1] S[table-format=4.1] S[table-format=3.1]}
\toprule
& {\textbf{Overall}} & \multicolumn{6}{c}{\textbf{Feasibility by scheduling family} (\%)} & \multicolumn{2}{c}{\textbf{Gap on feasible} (\%)} \\
\cmidrule(lr){3-8} \cmidrule(lr){9-10}
\textbf{Model} & {\textbf{Feas.\ \%}} & {ITC} & {INRC} & {NSPLib} & {JSPLib} & {RCPSP-MM} & {RCPSP-SM} & {Mean} & {Median} \\
                & & {($n_d{=}21$)} & {($n_d{=}69$)} & {($n_d{=}80$)} & {($n_d{=}100$)} & {($n_d{=}100$)} & {($n_d{=}100$)} & & \\
\midrule
gpt-5.4 (2026-03-05)                      & \bfseries 39.6 & \bfseries 0.0 & 44.9 & 0.0 & \bfseries 100.0 & \bfseries 42.0 & 13.0 & 139.6 & 3.2 \\
qwen/qwen3.5-397b (2026-02-16)$^{*}$      & 27.5 & \bfseries 0.0 & 30.9 & 8.1 & 51.0 & 32.0 & \bfseries 17.0 & 109.4 & \bfseries 0.0 \\
gpt-5.4-mini (2026-03-17)                 & 27.0 & \bfseries 0.0 & 21.7 & 2.5 & 96.0 & 14.0 & 0.0 & 146.5 & 9.4 \\
gemini-3.1-flash-lite                     & 20.6 & \bfseries 0.0 & \bfseries 65.2 & 1.2 & 43.0 & 0.0 & 8.0 & 529.3 & 23.7 \\
qwen/qwen3.5-122b (2026-02-24)$^{*}$      & 19.4 & \bfseries 0.0 & 18.5 & \bfseries 10.1 & 61.0 & 4.0 & 5.0 & 124.8 & 2.0 \\
qwen/qwen3.5-27b (2026-02-24)$^{*}$       & 14.6 & \bfseries 0.0 & 1.5 & 1.3 & 53.0 & 5.0 & 7.1 & \bfseries 31.8 & 2.6 \\
claude-haiku-4-5 (2025-10-01)             & 6.0 & \bfseries 0.0 & 26.1 & 5.0 & 6.0 & 0.0 & 0.0 & 771.8 & 176.9 \\
meta-llama-4-maverick-17b-123e            & 1.3 & \bfseries 0.0 & 4.3 & 0.0 & 2.0 & 0.0 & 1.0 & 311.3 & 205.9 \\
meta-llama-3.3-70b                        & 0.6 & \bfseries 0.0 & 1.4 & 0.0 & 2.0 & 0.0 & 0.0 & 3126.1 & 3369.2 \\
\midrule
\itshape Average across models   & \itshape 17.4 & \itshape 0.0 & \itshape 23.9 & \itshape 3.1 & \itshape 46.0 & \itshape 10.8 & \itshape 5.7 & \itshape 587.9 & \itshape 421.4 \\
\bottomrule
\end{tabular}%
}
\caption{Per-model \textbf{solver-code results on the v4 (full-variation)} variant of \textsc{SchedBench} (all three surface-form axes: themed entity names, paraphrased templates, and constraint reordering, active simultaneously). Feasibility rate counts instances where sandbox-executed model code produced a schedule with zero constraint violations; mean and median percent optimality gap are computed over feasible outputs only ($(z - z^{\star})/z^{\star} \times 100$). Higher feasibility is better; lower gap is better. The best value in each column is shown in \textbf{bold}. Per-family instance counts $n_d$ are listed below the column headers; $n=470$ total.}
\label{tab:solver-v4-470-full}
\end{table*}

\begin{table*}[t]
\centering
\small
\sisetup{table-number-alignment=center, detect-weight=true, detect-inline-weight=math}
\resizebox{\textwidth}{!}{%
\begin{tabular}{l S[table-format=3] S[table-format=1.3] l l l l c}
\toprule
\multicolumn{8}{c}{\textbf{v1 (plain) $\leftrightarrow$ v3A (themed) --- domain-framing axis}} \\
\midrule
\textbf{Model} & {$N^{\dagger}$} & {v1 viol.\ rate} & {$\bm{|\Delta\text{viol.\ rate}|}$} & {$\bm{|\Delta\text{coverage}|}$} & {Seed band $|\Delta\mathrm{vr}|$} & {95\% CI} & {Beyond noise?} \\
                & {\scriptsize paired} & {\scriptsize median} & {\scriptsize \textbf{median [Q1,\,Q3]}} & {\scriptsize \textbf{median [Q1,\,Q3]}} & {\scriptsize median [Q1,\,Q3]} & {\scriptsize $\Delta$med (eff$-$noise)} & \\
\midrule
meta-llama-4-maverick-17b-123e            & 470 & 0.272 & \textbf{0.076} {\scriptsize [0.025,\,0.167]} & 0.000 {\scriptsize [0.000,\,0.067]} & 0.092 {\scriptsize [0.025,\,0.269]} & [-0.042,\,-0.001] & no {\scriptsize ($<$ noise)} \\
meta-llama-3.3-70b                        & 470 & 0.214 & \textbf{0.066} {\scriptsize [0.006,\,0.199]} & 0.002 {\scriptsize [0.000,\,0.050]} & 0.087 {\scriptsize [0.020,\,0.295]} & [-0.045,\,+0.005] & near \\
qwen/qwen3.5-122b (2026-02-24)            & 467 & 0.080 & \textbf{0.062} {\scriptsize [0.007,\,0.182]} & 0.000 {\scriptsize [0.000,\,0.000]} & 0.114 {\scriptsize [0.031,\,0.369]} & [-0.089,\,-0.026] & no {\scriptsize ($<$ noise)} \\
qwen/qwen3.5-27b (2026-02-24)             & 470 & 0.045 & \textbf{0.048} {\scriptsize [0.000,\,0.180]} & 0.000 {\scriptsize [0.000,\,0.000]} & 0.091 {\scriptsize [0.005,\,0.335]} & [-0.069,\,-0.015] & no {\scriptsize ($<$ noise)} \\
claude-haiku-4-5 (2025-10-01)             & 470 & 0.315 & \textbf{0.033} {\scriptsize [0.006,\,0.128]} & 0.000 {\scriptsize [0.000,\,0.000]} & 0.089 {\scriptsize [0.014,\,0.219]} & [-0.076,\,-0.036] & no {\scriptsize ($<$ noise)} \\
claude-sonnet-4-6                         & 470 & 0.132 & \textbf{0.032} {\scriptsize [0.003,\,0.098]} & 0.000 {\scriptsize [0.000,\,0.058]} & 0.056 {\scriptsize [0.007,\,0.167]} & [-0.036,\,-0.006] & no {\scriptsize ($<$ noise)} \\
claude-opus-4-6                           & 468 & 0.074 & \textbf{0.031} {\scriptsize [0.000,\,0.060]} & 0.000 {\scriptsize [0.000,\,0.000]} & 0.031 {\scriptsize [0.000,\,0.083]} & [-0.005,\,+0.007] & near \\
gpt-5.4-mini (2026-03-17)                 & 450 & 0.240 & \textbf{0.031} {\scriptsize [0.006,\,0.136]} & 0.000 {\scriptsize [0.000,\,0.000]} & 0.065 {\scriptsize [0.007,\,0.200]} & [-0.056,\,-0.016] & no {\scriptsize ($<$ noise)} \\
qwen/qwen3.5-397b (2026-02-16)            & 470 & 0.033 & \textbf{0.031} {\scriptsize [0.000,\,0.097]} & 0.000 {\scriptsize [0.000,\,0.000]} & 0.047 {\scriptsize [0.000,\,0.206]} & [-0.039,\,-0.002] & no {\scriptsize ($<$ noise)} \\
gemini-3.1-flash-lite                     & 468 & 0.143 & \textbf{0.030} {\scriptsize [0.006,\,0.065]} & 0.000 {\scriptsize [0.000,\,0.002]} & 0.036 {\scriptsize [0.006,\,0.121]} & [-0.022,\,-0.001] & no {\scriptsize ($<$ noise)} \\
gpt-5.4 (2026-03-05)                      & 445 & 0.143 & \textbf{0.017} {\scriptsize [0.000,\,0.060]} & 0.000 {\scriptsize [0.000,\,0.000]} & 0.031 {\scriptsize [0.000,\,0.091]} & [-0.016,\,-0.005] & no {\scriptsize ($<$ noise)} \\
gemini-3-flash-preview                    & 469 & 0.000 & \textbf{0.010} {\scriptsize [0.000,\,0.087]} & 0.000 {\scriptsize [0.000,\,0.013]} & 0.033 {\scriptsize [0.000,\,0.175]} & [-0.041,\,-0.015] & no {\scriptsize ($<$ noise)} \\
gpt-5.5 (2026-04-23)                      & 470 & 0.000 & \textbf{0.000} {\scriptsize [0.000,\,0.000]} & 0.000 {\scriptsize [0.000,\,0.000]} & 0.000 {\scriptsize [0.000,\,0.033]} & [+0.000,\,+0.000] & near \\
\midrule
\itshape Median across models              & 470 & 0.132 & \itshape \textbf{0.031} & \itshape 0.000 & --- & --- & --- \\
\bottomrule
\end{tabular}%
}
\caption{Per-instance volatility under \textbf{v3A (themed)} on the 470-instance ablation set. $N^{\dagger}$ counts paired instances excluding no-output drops. The $|\Delta\text{viol.\ rate}|$ column reports absolute per-instance change; v1 violation rate serves as a capability anchor, since small changes can also indicate models are ``stably bad.'' The seed-band columns compare the effect against reseeding noise, with the bootstrap 95\% CI on $\mathrm{median}(|\text{effect}|)-\mathrm{median}(|\text{noise}|)$ determining the \emph{Beyond noise?} verdict. Verdicts: \emph{yes}: CI strictly $>0$. \emph{no ($<$ noise)}: CI strictly $<0$. \emph{near}: CI straddles 0 with lower bound above $-$IQR of the seed band. \emph{no}: effect and seed band both identically zero (no effect, no measurable noise).}
\label{tab:variation-v3a}
\end{table*}

\begin{table*}[t]
\centering
\small
\sisetup{table-number-alignment=center, detect-weight=true, detect-inline-weight=math}
\resizebox{\textwidth}{!}{%
\begin{tabular}{l S[table-format=3] S[table-format=1.3] l l l l c}
\toprule
\multicolumn{8}{c}{\textbf{v1 (plain) $\leftrightarrow$ v3B (paraphrased)}} \\
\midrule
\textbf{Model} & {$N^{\dagger}$} & {v1 viol.\ rate} & {$\bm{|\Delta\text{viol.\ rate}|}$} & {$\bm{|\Delta\text{coverage}|}$} & {Seed band $|\Delta\mathrm{vr}|$} & {95\% CI} & {Beyond noise?} \\
                & {\scriptsize paired} & {\scriptsize median} & {\scriptsize \textbf{median [Q1,\,Q3]}} & {\scriptsize \textbf{median [Q1,\,Q3]}} & {\scriptsize median [Q1,\,Q3]} & {\scriptsize $\Delta$med (eff$-$noise)} & \\
\midrule
meta-llama-4-maverick-17b-123e            & 470 & 0.272 & \textbf{0.068} {\scriptsize [0.017,\,0.167]} & 0.000 {\scriptsize [0.000,\,0.100]} & 0.055 {\scriptsize [0.010,\,0.145]} & [+0.001,\,+0.034] & \textbf{yes} \\
qwen/qwen3.5-122b (2026-02-24)            & 468 & 0.080 & \textbf{0.062} {\scriptsize [0.008,\,0.166]} & 0.000 {\scriptsize [0.000,\,0.000]} & 0.062 {\scriptsize [0.000,\,0.177]} & [-0.018,\,+0.025] & near \\
qwen/qwen3.5-27b (2026-02-24)             & 470 & 0.045 & \textbf{0.057} {\scriptsize [0.000,\,0.169]} & 0.000 {\scriptsize [0.000,\,0.000]} & 0.033 {\scriptsize [0.000,\,0.148]} & [-0.002,\,+0.042] & near \\
meta-llama-3.3-70b                        & 470 & 0.214 & \textbf{0.050} {\scriptsize [0.006,\,0.137]} & 0.000 {\scriptsize [0.000,\,0.034]} & 0.031 {\scriptsize [0.005,\,0.108]} & [-0.001,\,+0.033] & near \\
claude-sonnet-4-6                         & 470 & 0.132 & \textbf{0.040} {\scriptsize [0.004,\,0.104]} & 0.000 {\scriptsize [0.000,\,0.057]} & 0.031 {\scriptsize [0.000,\,0.083]} & [-0.007,\,+0.016] & near \\
gemini-3.1-flash-lite                     & 468 & 0.143 & \textbf{0.031} {\scriptsize [0.006,\,0.097]} & 0.000 {\scriptsize [0.000,\,0.031]} & 0.027 {\scriptsize [0.006,\,0.064]} & [-0.002,\,+0.012] & near \\
qwen/qwen3.5-397b (2026-02-16)            & 470 & 0.033 & \textbf{0.031} {\scriptsize [0.000,\,0.109]} & 0.000 {\scriptsize [0.000,\,0.000]} & 0.023 {\scriptsize [0.000,\,0.083]} & [-0.004,\,+0.016] & near \\
claude-opus-4-6                           & 470 & 0.074 & \textbf{0.030} {\scriptsize [0.004,\,0.062]} & 0.000 {\scriptsize [0.000,\,0.000]} & 0.023 {\scriptsize [0.005,\,0.050]} & [-0.002,\,+0.012] & near \\
claude-haiku-4-5 (2025-10-01)             & 470 & 0.315 & \textbf{0.025} {\scriptsize [0.000,\,0.094]} & 0.000 {\scriptsize [0.000,\,0.000]} & 0.017 {\scriptsize [0.000,\,0.091]} & [-0.011,\,+0.020] & near \\
gpt-5.4 (2026-03-05)                      & 446 & 0.143 & \textbf{0.023} {\scriptsize [0.005,\,0.062]} & 0.000 {\scriptsize [0.000,\,0.000]} & 0.021 {\scriptsize [0.001,\,0.062]} & [-0.005,\,+0.011] & near \\
gpt-5.4-mini (2026-03-17)                 & 445 & 0.240 & \textbf{0.023} {\scriptsize [0.001,\,0.091]} & 0.000 {\scriptsize [0.000,\,0.000]} & 0.017 {\scriptsize [0.001,\,0.074]} & [-0.005,\,+0.014] & near \\
gemini-3-flash-preview                    & 469 & 0.000 & \textbf{0.011} {\scriptsize [0.000,\,0.062]} & 0.000 {\scriptsize [0.000,\,0.000]} & 0.012 {\scriptsize [0.000,\,0.059]} & [-0.010,\,+0.009] & near \\
gpt-5.5 (2026-04-23)                      & 470 & 0.000 & \textbf{0.000} {\scriptsize [0.000,\,0.011]} & 0.000 {\scriptsize [0.000,\,0.000]} & 0.000 {\scriptsize [0.000,\,0.011]} & [+0.000,\,+0.000] & near \\
\midrule
\itshape Median across models              & 470 & 0.132 & \itshape \textbf{0.031} & \itshape 0.000 & --- & --- & --- \\
\bottomrule
\end{tabular}%
}
\caption{Per-instance volatility under \textbf{v3B (paraphrased)} on the 470-instance ablation set. $N^{\dagger}$ counts paired instances excluding no-output drops. The $|\Delta\text{viol.\ rate}|$ column reports absolute per-instance change; v1 violation rate serves as a capability anchor, since small changes can also indicate models are ``stably bad.'' The seed-band columns compare the effect against reseeding noise, with the bootstrap 95\% CI on $\mathrm{median}(|\text{effect}|)-\mathrm{median}(|\text{noise}|)$ determining the \emph{Beyond noise?} verdict. Verdicts: \emph{yes}: CI strictly $>0$. \emph{no ($<$ noise)}: CI strictly $<0$. \emph{near}: CI straddles 0 with lower bound above $-$IQR of the seed band. \emph{no}: effect and seed band both identically zero (no effect, no measurable noise).}
\label{tab:variation-v3b}
\end{table*}

\begin{table}[t]
\centering
\footnotesize
\setlength{\tabcolsep}{4pt}
\begin{tabular}{@{}lrrr@{}}
\toprule
Model & Clean & Empty & Total \\
\midrule
claude-haiku-4-5 (2025-10-01) & 470 & 0 & 470 \\
claude-opus-4-6 & 470 & 0 & 470 \\
claude-sonnet-4-6 & 470 & 0 & 470 \\
gemini-3-flash-preview & 469 & 1 & 470 \\
gemini-3.1-flash-lite & 469 & 1 & 470 \\
gpt-5.4 (2026-03-05) & 447 & 23 & 470 \\
gpt-5.4-mini (2026-03-17) & 450 & 20 & 470 \\
gpt-5.5 (2026-04-23) & 470 & 0 & 470 \\
meta-llama-3.3-70b & 470 & 0 & 470 \\
meta-llama-4-maverick-17b-123e & 470 & 0 & 470 \\
qwen/qwen3.5-122b (2026-02-24) & 468 & 2 & 470 \\
qwen/qwen3.5-27b (2026-02-24) & 470 & 0 & 470 \\
qwen/qwen3.5-397b (2026-02-16) & 470 & 0 & 470 \\
\midrule
\textbf{Total} & \textbf{6063} & \textbf{47} & \textbf{6110} \\
\bottomrule
\end{tabular}
\caption{Extraction success on \textbf{v1 (plain)} (470 instances $\times$ 13 models = 6110 cells). \textbf{Clean}: model output was successfully parsed into a candidate schedule by the extractor. \textbf{Empty}: output unparseable (formatting violations, model deviation). Overall clean rate: 99.2\%.}
\label{tab:extraction-v1}
\end{table}

\begin{table}[t]
\centering
\footnotesize
\setlength{\tabcolsep}{4pt}
\begin{tabular}{@{}lrrr@{}}
\toprule
Model & Clean & Empty & Total \\
\midrule
claude-haiku-4-5 (2025-10-01) & 470 & 0 & 470 \\
claude-opus-4-6 & 467 & 3 & 470 \\
claude-sonnet-4-6 & 465 & 5 & 470 \\
gemini-3-flash-preview & 470 & 0 & 470 \\
gemini-3.1-flash-lite & 469 & 1 & 470 \\
gpt-5.4 (2026-03-05) & 460 & 10 & 470 \\
gpt-5.4-mini (2026-03-17) & 448 & 22 & 470 \\
gpt-5.5 (2026-04-23) & 470 & 0 & 470 \\
meta-llama-3.3-70b & 470 & 0 & 470 \\
meta-llama-4-maverick-17b-123e & 470 & 0 & 470 \\
qwen/qwen3.5-122b (2026-02-24) & 468 & 2 & 470 \\
qwen/qwen3.5-27b (2026-02-24) & 470 & 0 & 470 \\
qwen/qwen3.5-397b (2026-02-16) & 467 & 3 & 470 \\
\midrule
\textbf{Total} & \textbf{6064} & \textbf{46} & \textbf{6110} \\
\bottomrule
\end{tabular}
\caption{Extraction success on \textbf{v2 (constraint shuffle)} (470 instances $\times$ 13 models = 6110 cells). \textbf{Clean}: model output was successfully parsed into a candidate schedule by the extractor. \textbf{Empty}: output unparseable (formatting violations, model deviation). Overall clean rate: 99.2\%.}
\label{tab:extraction-v2}
\end{table}

\begin{table}[t]
\centering
\footnotesize
\setlength{\tabcolsep}{4pt}
\begin{tabular}{@{}lrrr@{}}
\toprule
Model & Clean & Empty & Total \\
\midrule
claude-haiku-4-5 (2025-10-01) & 470 & 0 & 470 \\
claude-opus-4-6 & 468 & 2 & 470 \\
claude-sonnet-4-6 & 470 & 0 & 470 \\
gemini-3-flash-preview & 470 & 0 & 470 \\
gemini-3.1-flash-lite & 469 & 1 & 470 \\
gpt-5.4 (2026-03-05) & 451 & 19 & 470 \\
gpt-5.4-mini (2026-03-17) & 451 & 19 & 470 \\
gpt-5.5 (2026-04-23) & 470 & 0 & 470 \\
meta-llama-3.3-70b & 470 & 0 & 470 \\
meta-llama-4-maverick-17b-123e & 470 & 0 & 470 \\
qwen/qwen3.5-122b (2026-02-24) & 469 & 1 & 470 \\
qwen/qwen3.5-27b (2026-02-24) & 470 & 0 & 470 \\
qwen/qwen3.5-397b (2026-02-16) & 470 & 0 & 470 \\
\midrule
\textbf{Total} & \textbf{6068} & \textbf{42} & \textbf{6110} \\
\bottomrule
\end{tabular}
\caption{Extraction success on \textbf{v3A (themed)} (470 instances $\times$ 13 models = 6110 cells). \textbf{Clean}: model output was successfully parsed into a candidate schedule by the extractor. \textbf{Empty}: output unparseable (formatting violations, model deviation). Overall clean rate: 99.3\%.}
\label{tab:extraction-v3a}
\end{table}

\begin{table}[t]
\centering
\footnotesize
\setlength{\tabcolsep}{4pt}
\begin{tabular}{@{}lrrr@{}}
\toprule
Model & Clean & Empty & Total \\
\midrule
claude-haiku-4-5 (2025-10-01) & 470 & 0 & 470 \\
claude-opus-4-6 & 470 & 0 & 470 \\
claude-sonnet-4-6 & 470 & 0 & 470 \\
gemini-3-flash-preview & 470 & 0 & 470 \\
gemini-3.1-flash-lite & 469 & 1 & 470 \\
gpt-5.4 (2026-03-05) & 458 & 12 & 470 \\
gpt-5.4-mini (2026-03-17) & 448 & 22 & 470 \\
gpt-5.5 (2026-04-23) & 470 & 0 & 470 \\
meta-llama-3.3-70b & 470 & 0 & 470 \\
meta-llama-4-maverick-17b-123e & 470 & 0 & 470 \\
qwen/qwen3.5-122b (2026-02-24) & 470 & 0 & 470 \\
qwen/qwen3.5-27b (2026-02-24) & 470 & 0 & 470 \\
qwen/qwen3.5-397b (2026-02-16) & 470 & 0 & 470 \\
\midrule
\textbf{Total} & \textbf{6075} & \textbf{35} & \textbf{6110} \\
\bottomrule
\end{tabular}
\caption{Extraction success on \textbf{v3B (paraphrased)} (470 instances $\times$ 13 models = 6110 cells). \textbf{Clean}: model output was successfully parsed into a candidate schedule by the extractor. \textbf{Empty}: output unparseable (formatting violations, model deviation). Overall clean rate: 99.4\%.}
\label{tab:extraction-v3b}
\end{table}

\begin{table}[t]
\centering
\footnotesize
\setlength{\tabcolsep}{4pt}
\begin{tabular}{@{}lrrr@{}}
\toprule
Model & Clean & Empty & Total \\
\midrule
claude-haiku-4-5 (2025-10-01) & 1132 & 0 & 1132 \\
claude-opus-4-6 & 1129 & 3 & 1132 \\
claude-sonnet-4-6 & 1114 & 18 & 1132 \\
gemini-3-flash-preview & 1132 & 0 & 1132 \\
gemini-3.1-flash-lite & 1132 & 0 & 1132 \\
gpt-5.4 (2026-03-05) & 1120 & 12 & 1132 \\
gpt-5.4-mini (2026-03-17) & 1114 & 18 & 1132 \\
gpt-5.5 (2026-04-23) & 1132 & 0 & 1132 \\
meta-llama-3.3-70b & 1132 & 0 & 1132 \\
meta-llama-4-maverick-17b-123e & 1132 & 0 & 1132 \\
qwen/qwen3.5-122b (2026-02-24) & 1129 & 3 & 1132 \\
qwen/qwen3.5-27b (2026-02-24) & 1132 & 0 & 1132 \\
qwen/qwen3.5-397b (2026-02-16) & 1130 & 2 & 1132 \\
\midrule
\textbf{Total} & \textbf{14660} & \textbf{56} & \textbf{14716} \\
\bottomrule
\end{tabular}
\caption{Extraction success on \textbf{v4 (full-variation)} (1132 instances $\times$ 13 models = 14716 cells). \textbf{Clean}: model output was successfully parsed into a candidate schedule by the extractor. \textbf{Empty}: output unparseable (formatting violations, model deviation). Overall clean rate: 99.6\%.}
\label{tab:extraction-v4}
\end{table}

\clearpage
\begin{figure*}[!ht]
\centering
\footnotesize
\framebox{\begin{minipage}{0.95\textwidth}
\raggedright

\textbf{SchedBench instance \texttt{la04}} \hfill JSSP, 10 jobs $\times$ 5 machines,
BKS $=590$ \\
\emph{v4 rendering (themed entity names, paraphrased templates, constraint reordering)}

\vspace{4pt}\hrule\vspace{4pt}

In this scheduling problem, each named item consists of a sequence of numbered steps that must be performed in the exact order given.
Each step requires a specific named location and takes a fixed amount of time.

\par\smallskip
Your task is to produce a feasible schedule that satisfies all constraints and minimizes the total completion time (makespan).

\par\smallskip
Rules:

\par\smallskip
- Steps within the same item must follow their given order.
- Step k+1 for an item cannot start until step k has completed.
- A location can process at most one step at any given time.
- Steps cannot be interrupted once they start and must run to completion (non-preemptive).

\par\smallskip
Where:
- Start times are non-negative integers.

\par\smallskip
There are 10 scenes and 5 crew stations. In canonical order, the scenes are Scene Franklin, Scene Carlton, Scene Grant, Scene Pleasant, Scene Millbrook, Scene Lakeview, Scene Springfield, Scene Oakwood, Scene Kingston, and Scene Westport. The crew stations are North Sound Stage, Stage A Screening Room, Second Unit Screening Room, Practical Color Suite, and Stage B Editing Suite. Each scene has an ordered list of operations to perform; each lists a crew station and a duration.

\par\smallskip
Scene Franklin has a 5-step sequence. Steps 1-5: Step 1 at the North Sound Stage for 12 hours; Step 2 at the Second Unit Screening Room for 94 hours; Step 3 at the Practical Color Suite for 92 hours; Step 4 uses the Stage B Editing Suite for 91 hours; Step 5 uses the Stage A Screening Room for 7 hours.
Scene Carlton has a 5-step sequence. Steps 1-5: Step 1 uses the Stage A Screening Room for 19 hours; Step 2 uses the Practical Color Suite for 11 hours; Step 3 on the Stage B Editing Suite for 66 hours; Step 4 on the Second Unit Screening Room for 21 hours; Step 5 at the North Sound Stage for 87 hours.
Scene Westport: steps in order. Steps 1-5: Step 1 at the Second Unit Screening Room for 54 hours; Step 2 uses the Stage B Editing Suite for 5 hours; Step 3 at the Practical Color Suite for 59 hours; Step 4 uses the Stage A Screening Room for 15 hours; Step 5 on the North Sound Stage for 88 hours.
Scene Grant has an ordered sequence. Step 1 runs on the Stage A Screening Room for 14 hours. Step 2 is at the North Sound Stage for 75 hours. Step 3 is at the Practical Color Suite for 13 hours. Step 4 runs on the Stage B Editing Suite for 16 hours. Step 5 uses the Second Unit Screening Room for 20 hours.
Scene Lakeview follows an ordered sequence. Step 1 uses the Practical Color Suite for 77 hours. Step 2 is at the Second Unit Screening Room for 20 hours. Step 3 is at the North Sound Stage for 76 hours. Step 4 is at the Stage B Editing Suite for 88 hours. Step 5 uses the Stage A Screening Room for 53 hours.
Scene Oakwood has an ordered sequence. Step 1 uses the Stage A Screening Room for 88 hours. Step 2 uses the Practical Color Suite for 69 hours. Step 3 is at the North Sound Stage for 62 hours. Step 4 runs on the Stage B Editing Suite for 98 hours. Step 5 uses the Second Unit Screening Room for 52 hours.
Scene Pleasant: ordered steps. Step 1 uses the Second Unit Screening Room for 95 hours. Step 2 uses the Stage B Editing Suite for 66 hours. Step 3 runs on the North Sound Stage for 7 hours. Step 4 is at the Practical Color Suite for 7 hours. Step 5 uses the Stage A Screening Room for 77 hours.
Scene Kingston has an ordered sequence. Step 1 uses the Second Unit Screening Room for 61 hours. Step 2 runs on the Stage B Editing Suite for 9 hours. Step 3 runs on the North Sound Stage for 62 hours. Step 4 uses the Stage A Screening Room for 52 hours. Step 5 runs on the Practical Color Suite for 90 hours.
Scene Millbrook requires ordered steps. Step order: Step 1 on the Stage A Screening Room for 45 hours; Step 2 at the Practical Color Suite for 6 hours; Step 3 at the Stage B Editing Suite for 89 hours; Step 4 uses the North Sound Stage for 15 hours; Step 5 on the Second Unit Screening Room for 34 hours.
Scene Springfield: ordered steps. Step 1 is at the Second Unit Screening Room for 74 hours. Step 2 uses the Stage A Screening Room for 88 hours. Step 3 is at the North Sound Stage for 52 hours. Step 4 runs on the Practical Color Suite for 27 hours. Step 5 runs on the Stage B Editing Suite for 9 hours.

\vspace{4pt}\hrule\vspace{4pt}
\textbf{Response format:}

\vspace{2pt}
Response Format:
\textless{}ItemName\textgreater{} step 1: start=\textless{}integer\textgreater{}
\textless{}ItemName\textgreater{} step 2: start=\textless{}integer\textgreater{}
...

\par\smallskip
Example: "Shipment Madison step 1: start=0"

\par\smallskip
Rules:
- Item names must match the names used in the instance text exactly.
- Item names may contain spaces (e.g., "Shipment Madison"); reproduce them verbatim, including spacing and capitalization.
- Step numbering is 1-based and matches the numbering used in the instance text.
- Output one line per step and include every step exactly once.
- Output schedule lines only; do not add a makespan line or any other text.

\end{minipage}}
\caption{Complete example of a \textsc{SchedBench} instance under the \textbf{v4 (full-variation)} rendering. }
\label{fig:schedbench-instance-example}
\end{figure*}

\end{document}